\documentclass[journal]{IEEEtran}
\usepackage{amsmath,amsfonts}
\usepackage{algorithmic}
\usepackage{algorithm}
\usepackage{array}
\usepackage[caption=false,font=normalsize,labelfont=sf,textfont=sf]{subfig}
\usepackage{textcomp}
\usepackage{threeparttable}
\usepackage{stfloats}
\usepackage{url}
\usepackage{verbatim}
\usepackage{multirow}
\usepackage{graphicx}
\usepackage{cite}
\usepackage{listings}
\usepackage{fancyvrb}
\usepackage{framed}
\usepackage{hyperref}
\hypersetup{
    colorlinks=true,
    linkcolor=magenta,
    filecolor=magenta,      
    urlcolor=black,
    pdftitle={FINDER},
    pdfpagemode=FullScreen,
    }
\usepackage[english]{babel}

\usepackage[listings,skins]{tcolorbox}
\usepackage[thinlines]{easytable}
\usepackage[skipbelow=\topskip,skipabove=\topskip]{mdframed}
\mdfsetup{roundcorner=0}
\usepackage{xparse}
\usepackage[]{listings}
\usepackage{xcolor}
\usepackage{color,soul}
\usepackage{amsfonts}
\usepackage{amsmath}
\usepackage{amssymb}
\usepackage{bm}
\usepackage{textcomp,mathcomp}
\sethlcolor{yellow}
\usepackage{soul}
\usepackage{bbding}
\newcommand{\algorithmfootnote}[2][\footnotesize]{%
  \let\old@algocf@finish\@algocf@finish% Store algorithm finish macro
  \def\@algocf@finish{\old@algocf@finish% Update finish macro to insert "footnote"
    \leavevmode\rlap{\begin{minipage}{\linewidth}
    #1#2
    \end{minipage}}%
  }%
}
\makeatother

\newcolumntype{P}[1]{>{\centering\arraybackslash}p{#1}}
\newcolumntype{M}[1]{>{\centering\arraybackslash}m{#1}}

\makeatletter
\def\lst@makecaption{%
  \def\@captype{table}%
  \@makecaption
}
\makeatother
\begin{document}

\title{FINDER: Stochastic Mirroring of Noisy Quasi-Newton Search and Deep Network Training}

    “This work has been submitted to the IEEE for possible publication. Copyright may be transferred without notice, after which this version may no longer be accessible.”

\author{\IEEEauthorblockN{Uttam Suman\IEEEauthorrefmark{1},  Mariya Mamajiwala\IEEEauthorrefmark{2}, Mukul Saxena\IEEEauthorrefmark{1}, 
Ankit Tyagi\IEEEauthorrefmark{1}, Debasish Roy\IEEEauthorrefmark{1}\IEEEauthorrefmark{3}}\\
\IEEEauthorrefmark{1} Computational Mechanics Lab, Department of Civil Engineering, Indian Institute of Science Bangalore, Karnataka 560012, India\\
\IEEEauthorrefmark{2} School of Mathematical Sciences, University of Nottingham, University Park, Nottingham, NG7 2RD, UK\\
\IEEEauthorrefmark{3}{Corresponding author: royd@iisc.ac.in}}
\maketitle
\begin{abstract}
% \st{We propose a noise-assisted quasi-Newton optimizer and show its applications to neural network training. Being noise-assisted, our approach effectively surpasses local minima traps and addresses the instabilities caused by vanishing and exploding gradients, a common issue in training deep or wide neural networks. Furthermore, the quasi-Newton character makes the optimizer competitive against the state-of-the-art optimization algorithms like Adam and L-BFGS.}

We use a stochastic route to develop a new optimizer that has features suitable for training wide and deep neural networks. Optimization of such networks is hindered by the presence of multiple local minima and the instability caused by vanishing or exploding gradients, which can slow down the training process and increase the risk of convergence to a suboptimal solution. The proposed optimizer, acronymed FINDER (Filtering Informed Newton-like and Derivative-free Evolutionary Recursion), bridges noise-assisted global search and faster local convergence, the latter being a feature of quasi-Newton search using the principles of nonlinear stochastic filtering to construct a gain matrix that mimics the inverse Hessian in Newton's method. Following certain simplifications of the update to enable linear scaling with dimension and a few other enhancements, we first test the performance of FINDER against a few benchmark objective functions. Then we apply FINDER to training several Physics-Informed Neural Networks (PINNs), which involve minimizing highly nonlinear and non-convex loss functions. We consider networks with different widths and depths and compare the performance of FINDER with L-BFGS and Adam optimizers. Although we generally assume a deterministic loss function, we also address the modifications needed for noisy loss functions and demonstrate the effectiveness of FINDER through a pilot study on the MNIST classification problem.

\end{abstract}


\section{Introduction}
\IEEEPARstart{O}{ptimization} is ubiquitous. By systematically identifying the bespoke solution within a feasible set, optimization facilitates informed decision making and efficient routes to the desired outcomes. Our goal here is to extremize a real-valued objective or loss function $f$ typically arising in the training of deep neural nets (DNNs). The landscape of optimization methods may be broadly divided into deterministic and stochastic. Prominent among the former are linear and nonlinear programming \cite{luenberger1984linear,boyd2004convex} as well as variants of gradient descent \cite{ruder2017overview}. 
While effective in cases where $f$ is convex, these methods face challenges when applied to non-convex problems \cite{afshar2010size,faramarzi2014novel}. Newton's methods utilize the Hessian of $f$ \cite{bertsekas1982projected,ford1994multi,thapa1983optimization, izmailov2014newton,polyak2007newton} and enjoy quadratic convergence, thus marking a great leap in deterministic convex optimization. However, the risk of being trapped in a local optimum is significant in such cases. Additionally, inverting the Hessian matrix in higher dimensions demands high computational overhead (even with lower-rank approximations of the inverse). Other limitations include possible sensitivity to the initial guess and a risk of divergence in the presence of non-convexity or non-smoothness. Real-world problems are usually non-convex, an instance being the training of Physics-Informed Neural Networks (PINNs) \cite{raissi2017physics, haghighat2021physics}. Addressing this challenge requires carefully crafted strategies that navigate multimodality whilst not sacrificing the convergence rate. 

Being noise driven and thus enabled with global search for non-convex problems, stochastic methods have attracted attention  \cite{robbins1951stochastic}. A special note may be taken of meta-heuristic algorithms, where noisy search is often inspired by biological evolutions or some other physical or social phenomena.
Unlike derivative-based methods, meta-heuristics operate based only on the objective function values. 
Though generally lacking in mathematical rigor, they have often worked well in navigating complex search spaces. 
Some notable examples of meta-heuristics include Genetic Algorithms (GA) \cite{srinivas1994genetic}, Particle Swarm Optimization (PSO) \cite{wang2018particle} etc. 
However, with increasing dimension of the search space, the ensemble size to be sampled may grow exponentially and render such schemes ineffective. A few other derivative-free stochastic techniques are
Covariance Matrix Adaptation Evolutionary Strategy (CMA-ES) \cite{hansen2006cma}, simulated annealing \cite{rutenbar1989simulated} and differential evolution \cite{storn1997differential} etc.  
%\st{CMA-ES, in particular, stands out for its effectiveness in optimizing multimodal and non-smooth functions across myriad problem classes. Nevertheless, it faces drawbacks, including substantial computational overhead in higher dimensions and the presence of many hyper-parameters in the algorithm.} We refer to \cite{roy2024elements} for a comprehensive account of both deterministic and stochastic optimization methods.

For optimization in large dimensions, Adam \cite{kingma2014adam} and L-BFGS\cite{nocedal1980updating} represent the state-of-the-art -- the former dominating the class of Hessian-free first order optimizers while the latter leading quasi-Newton methods. Adam's key strength lies in handling noisy objective functions. It employs adaptive and dynamic step-size control based on past gradients, making it suitable for machine learning tasks. Being momentum driven, it manages to overcome local traps in the loss landscape. On the other hand, L-BFGS leverages a limited number of past gradients to approximate inverse Hessian and thus tackle high-dimensional problems. Yet, it has two bottlenecks - susceptibility to local minima and inability to handle stochastic objectives. When it comes to training wide and deep networks, performance of both optimizers may be worsened by multimodal loss functions and exploding or vanishing gradients. One way to tackle this difficulty could be by coupling the quasi-Newton approach with stochastic search whilst retaing the scaling capability. Such an optimization method, bridging linear scaling over dimensions and a good performance in global search, remains largely elusive. It is to this cause that the present article is devoted. 

%\st{For large-dimensional optimization problems, the Adaptive Moment Estimation (Adam)} \cite{kingma2014adam} - \st{a gradient-based scheme} - \st{has emerged as a powerful tool, especially for machine learning tasks. Its key strength is in adaptive and dynamic step-size control for each component of the state vector based on past gradients, resulting in faster convergence. Although Adam has advanced the state-of-the-art in optimization, it is not without limitations - the requirement of tuning several hyperparameters and possible convergence to sub-optimal solutions in non-convex and noisy problems, to name a few. Indeed, an optimization method, with twin abilities of linear scaling across dimensions and good performance at global search, remains largely elusive. It is to this cause that the present article is devoted. }

We draw upon the extensive machinery of stochastic calculus to craft a novel optimization method, acronymed FINDER. Drawing inspiration from measure-theoretic ideas in stochastic filtering, we show that an imposition of the constraint of vanishing gradient of $f$ through a so-called measurement equation yields a gain matrix (a stochastic equivalent of matrix-valued Lagrange multipliers) that may be interpreted as the mimicked inverse Hessian. This circumvents an explicit computation of the Hessian and its inverse. A grounding within the theory of diffusive stochastic processes ensures that the capability for noise-assisted exploration is never sacrificed. We expect FINDER, being quasi-Newton as well as noise-driven, to display its strength in application in optimizing wide and deep neural networks.

The rest of the paper is organized as follows. Section \ref{s2} presents the broad mathematical framework for posing the task of optimization as a nonlinear filtering problem. This section also includes a detailed algorithm for the proposed optimizer. Section \ref{s3} demonstrates the numerical performance of the optimizer against a set of IEEE benchmark functions, followed by the training of deep and wide neural networks typically arising in PINNs. Finally, some concluding remarks are given in Section \ref{s4}. Table \ref{T1} gives an outline of notations used in this article. The distinction between scalar and vector quantities should be apparent from the context.

\begin{table}
\centering
\caption{Table of Notations}\label{T1}
\begin{tabular}{|c||c|}
\hline
\textbf{Notation} & \textbf{Description} \\
\hline
$\boldsymbol{A}$ & matrix\\
$A$ & a vector random variable\\
$t$ & time-like quantity \\
$A_t$ & a random (stochastic) process\\
$\mathcal{N}$ & normal probability distribution\\
$\mathcal{U}$ & uniform probability distribution\\
$A^{j}$ & $j^{th}$ realization of $A$ \\
$f$ & objective (loss) function\\
$\nabla f$ & gradient of $f$ (a column vector)\\
$\langle u,v\rangle$ & inner product of $u$ and $v$\\
$\langle u,v\rangle_{\boldsymbol{A}}$ & inner product of $u$ and $\boldsymbol{A}v$ \\
$\mathbb{E}_\mathbb{P}[A]$ & expectation of $A$ under measure $\mathbb{P}$\\
w.r.t. $a\uparrow$ & with respect to increasing order of vector $a$\\
argsort($a, \boldsymbol{A, ..}$) & sort columns of ($\boldsymbol{A, ..}$) w.r.t. $a\uparrow$\\
$\boldsymbol{A}^s$ & matrix after sorting columns in $\boldsymbol{A}$\\
$\overline{\boldsymbol{A}}$ & arithmetic mean of all columns vectors of $\boldsymbol{A}$\\
$\boldsymbol{A}^{j}$ & $j^{th}$ column vector in $\boldsymbol{A}$\\
$\boldsymbol{A}_{ij}$ & $(i,j)^{th}$ entry of $\boldsymbol{A}$\\
$[a]_{m\times n}$ & $m\times n$ matrix with every element being $a$\\
$\boldsymbol{I}_n$ & rank $n$ identity matrix\\
$u_{,x_1\ldots x_n}$ & $\frac{\partial^n u}{\partial x_1\ldots \partial x_n}$ \\
$\boldsymbol{A}^T$ & transpose of matrix $\boldsymbol{A}$\\
$a \wedge b$ & minimum of $a$ and $b$\\
\hline
\end{tabular}
\end{table}

\section{Mathematical Formulation}\label{s2}
\subsection{Problem Statement}
Consider the following optimization problem:
\begin{equation}
  \underset{x}{\operatorname{argmin}}\ f(x) \text{ where } x\in \text{D}\subseteq\mathbb{R}^{N}  
\end{equation}
However, for all practical purposes, we may recast the problem as follows:
$\text{given a cost function } f: \mathbb{R}^N\rightarrow\mathbb{R}^{+}\cup\{0\},$
$$\text{find } x^{*}\in\text{D}\subseteq\mathbb{R}^{N}\ \text{such that } f(x^{*})\leq\varepsilon_{tol}$$ where $\varepsilon_{tol}$ is a user-defined tolerance. 
Ideally, $f(x)$ should be differentiable w.r.t. $x$, enabling the use of gradient-based optimization techniques. In case $f$ is continuous but not uniformly differentiable over $\text{D}$, one may evaluate a stochastic variant of the gradient $\nabla{f}$ based on the proposal in \cite{saxena2022microstructure}. %Continuity and smoothness of $f$ are desirable attributes, aiding in convergence and stability during the optimization process.

We first pose the optimization problem from a filtering perspective. Stochastic filtering aims at estimating the hidden state of a system conditioned on certain noisy measurements. A parallel may be drawn with the optimization of a smooth function that requires finding the state vector $x^{*}$ conditioned on $\nabla f(x^{*}) = 0$ modulo a zero-mean noise. Drawing upon this analogy, we may consider $f(x)$ as a function of the hidden state $x$ that is in turn a stochastic (Markov) process. We may measure noisy gradients $\nabla f(x)$ and require them to approach the zero vector modulo noise with zero mean. The Kushner-Strotonovich filtering equation provides a means to reach this goal, i.e., to recursively update the hidden state such that $\nabla f(x(t))$ approaches the zero vector modulo a zero-mean noise asymptotically. 

To put all this in a formal setting, consider a complete probability space $(\Omega,\mathcal{F},\mathbb{P})$ \cite{oksendal2013stochastic} equipped with filtration $(\mathcal{F}_{t})_{t\geq0}$ derived from subsets of the sample space $\Omega$. We examine a vector-valued diffusion process $X_{t}:=X(t)$, adapted to $\mathcal{F}_t$ and varying over the time-like interval $t\in [0,T]$. We wish to arrive at a stochastic dynamical scheme for $X_t$ so that $\mathbb{E}_\mathbb{P}(X_t)\rightarrow x^*$ as $t\rightarrow\infty$. Here, $X_t=\{(X_t)_i :\ i=1,\ldots, N\}$, denotes a weak solution to the following stochastic differential equation (SDE) in $\mathbb{R}^N$, driven by an $N$-dimensional standard Brownian motion $B_t$:
\begin{equation}\label{e1}
    \begin{aligned}
        dX_t=\boldsymbol{R}_t dB_t
    \end{aligned}
\end{equation}
where $\boldsymbol{R}_t$ is the diffusion matrix. Eqn. \eqref{e1} is referred to as the process dynamics and $X_t$ is the hidden state or process variable. To impose a stochastic equivalent of the constraint $\nabla f =0$, we use the principle of stochastic filtering to require that $\nabla f$ be reduced to a zero-mean martingale, i.e. a stochastic process whose expected future value conditioned on the present is the present value itself. Accordingly, we supplement Eqn. \eqref{e1} with a constraint or measurement equation in terms of the 'measured variable' $Y_t\in\mathbb{R}^N$ such that the update for $X_t$ mimics the one in Newton's method. Specifically, we write the measurement equation in the algebraic form:
 \begin{equation}\label{e222}
     \begin{aligned}
         Y_t = \nabla f + \dot{W}_t
     \end{aligned}
 \end{equation}
and require $Y_t$ (and hence $\nabla{f}$) to be a zero-mean noise weakly, i.e. with posterior probability 1, for all $t$. Here $W_t \in \mathbb{R}^N$ denotes the measurement noise -- another Brownian motion independent of $B_t$, so that $\dot{W}_t$ is formally a white noise. Denoting the filtration generated by $Y_t$ by $\mathcal{F}_t^{Y}$, the estimate $\hat{X}_t$ of $X_t$ under this constraint is given by the conditional expectation $\mathbb{E}_\mathbb{P}(X_t|\mathcal{F}_t^{Y})$. Our aim is to find a computationally expedient, $t$-recursive update for $\hat{X}_t$.  

In nonlinear stochastic filtering \cite{van2007stochastic}, the aim of reducing $\nabla{f}$ to a zero-mean noise is achieved through a change of measures. In this setting, $I_t = \nabla f(X_t)$ is called the innovation vector and the change of measures leads to a generalized Bayes rule (the Kallianpur-Striebel formula) or, equivalently, the Kushner-Stratonovich equation to evolve $\hat{X}_t$ \cite{kushner1964differential,stratonovich1968conditional,sarkar2014kushner}. A simpler and approximate update, valid in principle for Gaussian noises and linear dynamics, is afforded by the ensemble Kalman filter (EnKF) \cite{saxena2024, roy2017stochastic}. For an interval $\Delta t$, a recursive update for $\hat{X}_t$ is given by:
\begin{equation}\label{e5}
 \begin{aligned}
         \hat{X}_{t+\Delta t}=\hat{X}_t+\boldsymbol{\Tilde{G}}I_{t} = \hat{X}_t+\boldsymbol{\Tilde{G}}\nabla f(\hat{X}_t) 
 \end{aligned}
 \end{equation}
Here $\boldsymbol{\Tilde{G}}$ is the gain matrix, a stochastic variant of the matrix of Lagrange multipliers to drive $\nabla{f}$ to a zero-mean noise (martingale). When this objective is met, we have $\mathbb{E}_\mathbb{P}(\hat{X}_t)\rightarrow x^*$. Upon a direct comparison with Newton's update, it is evident that $\boldsymbol{\Tilde{G}}$ serves as a stochastic mirror of the inverse Hessian. An explicit expression of $\boldsymbol{\Tilde{G}}$ is as follows:

\begin{equation}\label{G}
   \begin{aligned}
 \boldsymbol{\Tilde{G}}&=-\left[\int_\Omega \left(\boldsymbol{X}-\boldsymbol{\overline{X}}\right)\left(\boldsymbol{G}-\boldsymbol{\overline{G}}\right)^Td\mathbb{P}\right]\\&\left[\int_\Omega  \left(\boldsymbol{G}-\boldsymbol{\overline{G}}\right)\left(\boldsymbol{G}-\boldsymbol{\overline{G}}\right)^Td\mathbb{P} + \boldsymbol{Q}\right]^{-1}
     \end{aligned}
 \end{equation}
where $\boldsymbol{X} = [X_t^1, X_t^2,\ldots, X_t^p]\in \mathbb{R}^{N\times p}$ is the matrix of ensemble members or particles; $X_t^i$ is the $i^{\text{th}}$ particle and $\boldsymbol{G} = [\nabla f_t^1, \nabla f_t^2,\ldots, \nabla f_t^p]\in \mathbb{R}^{N\times p}$ is the matrix of $\nabla{f}(x)$ evaluated at each particle location. See the Supplementary Material \href{https://github.com/FINDER-optimizer/FINDER/blob/main/supplementary_material.pdf}{\underline{here}} for a derivation of the update. 
$\boldsymbol{Q}=\mathbb{E}_\mathbb{P} [dW_t dW_t^T]$ is the measurement noise covariance matrix; In practical Monte Carlo (MC) implementations involving finite (possibly small) ensembles, we take the Frobenius norm $||\boldsymbol{Q}||$ to be quite small.

We may obtain $\nabla f$ using automatic differentiation. However, an issue with the method above is that the matrix inverse appearing in $\boldsymbol{\Tilde{G}}$ prevents a linear scaling across dimensions. Another computational hurdle is with MC simulations of the equations involved, where the number of particles may need to increase exponentially with dimension in the absence of a trustworthy variance reduction strategy. These call for modifications and simplifications, which we now undertake.

\subsection{Implementation and Algorithm}
As in Section A, the algorithm begins with Eqn. \eqref{e1}, which is used to sample particles around a mean (e.g. the last available estimate). This is followed by evaluating gradients $\nabla f$ (Eqn. \eqref{e222}) at these particles. The particles and gradients obtain a mimicked inverse Hessian as in Eqn. \eqref{G}, which is used to update the ensemble via Eqn. \eqref{e5}. A step-by-step elucidation of the FINDER algorithm with code-snippets is available in the Supplementary Material. \textit{For a clear understanding of the algorithm, follow the walkthrough example in the Supplementary Material.}

\textbf{Step 1 -- Generate Ensemble:} From Eqn. \eqref{e1}, we have  $(dB_t)_i\sim\mathcal{N}(0,dt)\Rightarrow (\Delta B_{t})_i\sim\mathcal{N}(0,1)$ upon discretization with $\Delta t = 1$, where subscript $i$ denotes the $i^{th}$ component of the vector. We weakly replace this Gaussian increment with a zero-mean uniform density of spread $[-1,1]$ and a proper choice of the diagonal matrix $\boldsymbol{R}_t$ so as to match the second moments. Compared with a Gaussian distribution, such a choice leads to less sampling time, especially in higher dimensions. Accordingly, consider a vector random process $Z_t\in\mathbb{R}^N$ such that $Z_t\sim\prod_{i=1}^N\mathcal{U}(-1,1)$ for all $t$. Then we have:
\begin{equation}\label{sampling}
    X_t^{j} = X_t^1 + \boldsymbol{R}_t Z_t^{j},\hspace{0.5cm}j=2,\ldots ,p
\end{equation}
where $X_0^1$ is the initial solution which is iteratively updated and $Z_t^j$ is the $j^{th}$ realization of the $N$ dimensional random process $Z_t$. We construct the ensemble matrix $\boldsymbol{X}_{N\times p}$ with particles as column vectors.

\textbf{Step 2 -- Evaluate Gradients:} Having evaluated gradients for all particles by automatic differentiation, we store them as columns in the gradients matrix $\boldsymbol{G}_{N\times p}$. We also evaluate the cost of each particle, stored in cost vector $f_{1\times p}$. Subsequently, we sort columns in $\boldsymbol{X}$ and $\boldsymbol{G}$ in increasing order of $f$ to obtain the sorted ensemble matrix $\boldsymbol{X}^s$ and sorted gradients $\boldsymbol{G}^s$. Note that sorting the $p$-dimensional cost vector is not computationally demanding.

\textbf{Step 3 -- Mimicked Inverse Hessian:} The evaluation of $\boldsymbol{\tilde{G}}$ via Eqn. \eqref{G} is computationally expensive, even infeasible, in higher dimensions. 
Hence, for computational expedience including variance reduction, we approximate $\boldsymbol{\Tilde{G}}$ by a diagonal matrix $\boldsymbol{B}$  [see Section 5.7.1 in \cite{bertsekas2021data}] whose elements are given by:
\begin{equation}\label{G1}
    \begin{aligned}
     \boldsymbol{B}_{ii} = 
        \frac{\sum_{j=1}^{p}(\boldsymbol{X}^s_{ij} -(\overline{\boldsymbol{X}^s})_i)(\boldsymbol{G}^s_{ij} - (\overline{\boldsymbol{G}^s})_i)}{\sum_{j=1}^{p}(\boldsymbol{G}^s_{ij} - (\overline{\boldsymbol{G}^s})_i)(\boldsymbol{G}^s_{ij} - (\overline{\boldsymbol{G}^s})_i)}
    \end{aligned}
\end{equation}
where $\boldsymbol{X}^s_{ij}$ and $\boldsymbol{G}^s_{ij}$ denote the $(ij)^{th}$ element of the sorted ensemble matrix $\boldsymbol{X}^s$ and the sorted gradients matrix $\boldsymbol{G}^s$ respectively. $(\overline{\boldsymbol{X}^s})_i = \frac{1}{p}\sum_{j=1}^{p}\boldsymbol{X}^s_{ij}$ and $(\overline{\boldsymbol{G}^s})_i = \frac{1}{p}\sum_{j=1}^{p}\boldsymbol{G}^s_{ij}$ are the $i^{th}$ components of the associated mean column vector.
%Note that the second factor (the inverse matrix) in the expression of the gain $\boldsymbol{\Tilde{G}}$ is indeed diagonal in principle. This follows from the statistical independence of the scalar components of $W_t$. However, in a finite-ensemble Monte Carlo simulation, such a diagonal structure is not generally realized. 
Apart from relieving the computational burden of matrix inversion, this approximation has the added benefit of variance reduction by ignoring particle scatter due to the off-diagonal entries. Above all, it reduces the number of particles (we use $p=5$) needed for computing $\boldsymbol{B}$. In this context, one may, in principle, use an appropriate change of measures to achieve `perfect' variance reduction, i.e. simulation with just one particle \cite{milstein2002monte}. However, given the inherent circularity in the implementation of this principle, we do not attempt it here.
We enforce $\boldsymbol{B}$ to be positive semi-definite so that $\langle \boldsymbol{B}\nabla f,\nabla f\rangle \geq 0$. This implies that the direction given by $-\boldsymbol{B}\nabla f$ is still a direction of descent. Accordingly, we set any spurious negative entry in $\boldsymbol{B}$ to zero. This makes sense, given that for a non-degenerate vector-valued random variable, diagonal entries of the covariance matrix must be positive -- a condition that may be violated during MC simulations with just a few particles.
To compensate for the loss of accuracy that accrues due to the approximation of $\boldsymbol{\Tilde{G}}$ by a diagonal matrix, we introduce a parameter $\gamma$ as an exponent of $\boldsymbol{B}$, i.e. $\boldsymbol{\Tilde{B}}:=\boldsymbol{B}^\gamma$ where $0\leq\gamma\leq 1$ \cite{yao2020adahessian}. Substituting $\boldsymbol{\Tilde{G}}$ with $\boldsymbol{\Tilde{B}}$ in Eqn. \eqref{e5}, it is evident that $\gamma=0$ implies a gradient descent-like update while $\gamma=1$ (used in our numerical work) implies a BFGS-mimicking update. 
Such a substitution also provides a handle to regulate the step size driven by the curvature of the design space topology. An intuitive grasp of the connection between curvature (derived from the inverse Hessian) at a point and the optimal step size suggests that when curvature is substantial, only a small step is necessary. Hence we choose $0\leq\gamma\leq 1$.

\textbf{Step 4 -- Update Ensemble:} We modify the update Eqn. \eqref{e5} to replace the increment $\boldsymbol{\Tilde{G}}I_{t}$ with the increment matrix $\boldsymbol{\Delta}_{t+1}$ given by:
\begin{equation}\label{increment}
    \boldsymbol{\Delta}_{t+1} = \theta \boldsymbol{\Delta}_{t} + \boldsymbol{\Tilde{B}G}^s    
\end{equation} 
where $\theta\in [0,1)$ and $\boldsymbol{\Delta}_0 = [0]_{N\times p}$. Since $\Delta t=1$ and we start the recursion with $t=0$, $t$ is a non-negative integer. Thus, we update the ensemble as:
\begin{equation}\label{e52}
 \begin{aligned}
\boldsymbol{\hat{X}}=\boldsymbol{X}^s-\alpha_t\boldsymbol{\Delta}_{t+1}
\end{aligned}
\end{equation}
where $\alpha_t\in (0,0.1]$ is a step-size multiplier and $\hat{\boldsymbol{X}}$ the updated ensemble matrix. Eqn. \eqref{increment} represents a convolution of past increments against exponentially decaying weights, i.e. $\boldsymbol{\Delta}_t = \sum_{k=1}^t\theta^k\boldsymbol{\Delta}_{t-k}$. $0\leq\theta < 1$ ensures that the weights decay in a geometric progression. The convolution step should help accelerate the algorithm in regions with slowly varying gradients. For example, if past increments align in the same direction, it leads to an increase in the value of  $\boldsymbol{\Delta}_{t+1}$. $\theta$ close to 1 implies higher reliance on past increments, whereas a value near 0 suggests otherwise. Another crucial element is the computation of the step-size multiplier $\alpha_t$. An appropriate $\alpha_t$ in the dominant descent direction $-\boldsymbol{\Delta}_{t+1}^{1}$ may be obtained iteratively using the Armijo rule: \cite{Nocedal2006}.
\begin{equation}\label{eq9}
    \begin{aligned}
    f\left(({\boldsymbol{X}^s})^{1}-\alpha_t\boldsymbol{\Delta}_{t+1}^{1}\right) \leq f\left(({\boldsymbol{X}^s})^{1}\right)-c_{\alpha}\alpha_t \langle\boldsymbol{\Delta}_{t+1}^{1}, {(\boldsymbol{G}^s})^{1}\rangle
    \end{aligned}
\end{equation}
where $c_\alpha\in[10^{-3}, 10^{-2}]$. $(\boldsymbol{X}^s)^1$ denotes the first column in the sorted ensemble matrix $\boldsymbol{X}^s$; similar interpretations apply to $\boldsymbol{\Delta}_{t+1}^1$ and $(\boldsymbol{G}^s)^1$. 
%denote the first column of increment matrix and gradient matrix respectively. 
We include the best particle $(\boldsymbol{X}^s)^{1}$ 
of the last ensemble in the updated ensemble matrix $\boldsymbol{\hat{X}}$ to obtain the new ensemble matrix $\boldsymbol{X}'_{N\times (p+1)}$. We obtain the new cost vector $f'$ for the new ensemble by evaluating the cost of each particle in $\boldsymbol{X}'$ and subsequently finding the 'best' particle ${(\boldsymbol{X}'^s})^{1}$, and the 'worst' particle ${(\boldsymbol{X}'^s})^{p+1}$. ${(\boldsymbol{X}'^s})^{1}$ serves as the new mean $X^{1}_{t+1}$ for the next iteration. This ensures that the loss function is non-increasing as FINDER latches on to the best particle in the last iteration until it finds a better one in the next iteration either through sampling or update.

\textbf{Step 5 -- Update Diffusion Matrix:} We update $\boldsymbol{R}_t$ as:
\begin{equation}\label{eq10}
\begin{aligned}
\Gamma_{t+1}&= (1 - c_s)\Gamma_t + c_s\left(({\boldsymbol{X}'^s})^{p+1} - ({\boldsymbol{X}'^s})^{1}\right)\\
(\boldsymbol{R}_{t+1})_{_{ii}}&=\begin{cases}
		|(\Gamma_{t+1})_i| \wedge \zeta_1  \text{ (say } 10^{-2}) 	, & \text{if ($\Gamma_{t+1})_i \neq 0$}\\
            \zeta_2 \text{ (say } 10^{-2}), & \text{otherwise}
		 \end{cases}  
\end{aligned}
\end{equation}
where $c_s = 0.1$; $\Gamma_0 = [0]_{N\times 1}$ and $\boldsymbol{R}_0 = 0.1\boldsymbol{I}_{N}$. The vector $\Gamma$ takes into account the difference between  $(\boldsymbol{X}'^s)^{p+1}$ and $(\boldsymbol{X}'^s)^1$  with a suitable weight $(1-c_s)$ applied to the history of such updates. The idea here is to sample the $i^{th}$ component from the range  $|(\boldsymbol{X}'^s)^{p+1}_i - (\boldsymbol{X}'^s)^1_i|$ around $(\boldsymbol{X}'^s)^1_i$ in the next iteration. $c_s = 0.1$ attaches higher weight to the history so the search span does not vary abruptly.
Following this idea, we wish to choose $|(\Gamma_{t+1})_i|$ as $(\boldsymbol{R}_{t+1})_{ii}$. However, $\boldsymbol{R}_{t+1}$ must have a bounded spectral radius so that all particles are sampled from a bounded neighborhood around the mean. Accordingly, we prescribe an upper limit on $|(\Gamma_{t+1})_i|$, say $\zeta_1$ to obtain $(\boldsymbol{R}_{t+1})_{ii}$. It is also appropriate to avoid an unphysical zero value for $(\boldsymbol{R}_{t+1})_{ii}$ to avoid nil sampling along the $i^{th}$ component. The prescribed limits on the elements of $\boldsymbol{R}_{t+1}$ may themselves be treated as hyper-parameters ($\zeta_1$ and $\zeta_2$) if they aid in finding the minima with higher precision. As the estimation of $\boldsymbol{B}$ depends on the spread of the ensemble around the mean, the performance of FINDER is sensitive to the choice of $\zeta_1, \zeta_2$. Higher $\zeta_1,\zeta_2$ render the algorithm more exploratory. Diagonal values of $\boldsymbol{R}_{t+1}$ could be interpreted as the individual variances for zero-mean uniform densities. A weak replacement of the Gaussian density is admissible \cite{milstein2013numerical} as it does not interfere with the local rate of convergence of the Euler-Maruyama method for discretizing the process SDE.

\textbf{Pseudo-code:} The pseudo-code of the proposed optimization scheme is furnished in  Algorithm \ref{algo1}. 
\begin{algorithm}
 \caption{FINDER}\label{algo1}
\begin{align*}
&\textbf{Set } p=5,\theta=0.9, \gamma=1,c_s=0.1, c_\alpha=0.01, \\
&\quad\quad\zeta_1=\zeta_2=10^{-4}\quad\text{\# hyperparameters}\\
&\text{\textbf{Initialize} }X_{0}^1, \boldsymbol{R}_0 = 0.1I_{_N}, \boldsymbol{ \Delta}_0=[0]_{N\times p}, \Gamma_0=0, t=0\\
 &\text{\textbf{while} $f>\varepsilon_{tol}$ and $t\leq T$  \textbf{do:}} \\
&\quad \textbf{for }j = 2\text{ to }p\ \textbf{do}: \quad \text {\# generate ensemble}\\
&\quad \quad X^{j}_{t} = X_{t}^1 + \boldsymbol{R}_t\ Z_t^{j} \text{ where }Z_t\sim\prod_{i=1}^{N}\mathcal{U}(-1,1) \\
&\quad \textbf{end}\\
&\quad \boldsymbol{X} = [X_t^1, X_t^{2},\ldots X_t^{p}]_{N\times p}  \quad \text{\# ensemble matrix}\\
&\quad \boldsymbol{G} = [\nabla f(\boldsymbol{X})]_{N \times p} \quad \text{\# gradients matrix}\\
&\quad f = [f(\boldsymbol{X})]_{1\times p} \quad \text{\# cost vector for }\boldsymbol{X} \\
&\quad \boldsymbol{X}^s, \boldsymbol{G}^s \leftarrow \text{argsort}(f, \boldsymbol{X}, \boldsymbol{G}) \quad \text{\# Sort } \boldsymbol{X} \text{, } \boldsymbol{G} \text{ w.r.t. } f\uparrow\\
&\quad \text{Compute } \boldsymbol{\tilde{B}} = \boldsymbol{B}^{\gamma} \quad \text{\# use Eqn.\ref{G1} \& set $\boldsymbol{B}_{ii}$ = 0 if $\boldsymbol{B}_{ii}<$0 }\\
&\quad \boldsymbol{\Delta}_{t+1} = \theta \boldsymbol{\Delta}_t + \boldsymbol{\Tilde{B}G^s}\quad\text{\# compute increment matrix} \\
&\quad \alpha_t\leftarrow \text{Armijo rule}(({\boldsymbol{X}^s})^{1}, ({\boldsymbol{G}^s})^{1}, \boldsymbol{\Delta}^{1}_{t+1}, c_\alpha)\quad \text{\# Eqn. \eqref{eq9}}\\
&\quad \boldsymbol{\hat{X}} = \boldsymbol{X}^s - \alpha_t\boldsymbol{\Delta}_{t+1} \quad\text{\# update the ensemble }\\
&\quad \boldsymbol{X}' = [\boldsymbol{\hat{X}}, ({\boldsymbol{X}^s})^{1}]_{N\times (p+1)} \quad \text{\# concatenate $({\boldsymbol{X}^s})^{1}$ to $\boldsymbol{\hat{X}}$}\\
&\quad f' = [f(\boldsymbol{X}')]_{1\times (p+1)} \quad \text{ \# cost vector for $\boldsymbol{X}'$}\\
&\quad \boldsymbol{X}'^s\leftarrow\text{argsort(}f', \boldsymbol{X}'\text{)}\quad \text{\# sort }\boldsymbol{X}' \text{ w.r.t. } f'\uparrow\\
&\quad \Gamma_{t+1} = (1-c_s)\Gamma_t + c_s\left(({\boldsymbol{X}'^s})^{p+1} - ({\boldsymbol{X}'^s})^{1}\right)\\
&\quad \textbf{for }i = 1\text{ to }N\textbf{ do:}\\
&\quad \quad(\boldsymbol{R}_{t+1})_{_{ii}} = \begin{cases}
		|(\Gamma_{t+1})_i| \wedge \zeta_1   	, & \text{if ($\Gamma_{t+1})_i \neq 0$}\\
            \zeta_2 , & \text{otherwise}
		 \end{cases}\\
&\quad \textbf{end}\\
&\quad X^1_{t+1} \leftarrow ({\boldsymbol{X}'^s})^1\\
&\quad t = t + 1\\
&\text{\textbf{Return} $x^*$ = $X_t^1$}\quad \text{\# optimal solution } \\
&\text{\textbf{Output} } \text{Record and plot results for comparison.}
\end{align*}
\end{algorithm}

\subsection*{Remarks on convergence}

The adopted filtering framework follows the structure of EnKF \cite{evensen2003ensemble}, which applies to our case after linearization of Eqn. \eqref{e222}. As shown in \cite{schillings2017analysis}, the EnKF-based update map follows a gradient flow which we now briefly discuss. In doing so, we only focus on the broad framework of filtering-driven optimization, thereby largely bypassing the simplifications used for scaling with dimensions (e.g. diagonalization of $\boldsymbol{\Tilde{G}}$) and other performance-enhancing features (e.g. history dependence). Consider the simplified form of the update as in Eqn. \eqref{e5} along with the following measurement equation:
\begin{equation}
    0=\nabla f(X_t) + \dot{W}_t
\end{equation} 
The identity above holds almost surely. Central to our discussion is the minimization of the following functional:
\begin{equation}\label{eqn:Phi}
    \phi(x)=\frac{1}{2}||\boldsymbol{Q}^{-1/2}\nabla f(x)||^2
\end{equation}
where $\boldsymbol{Q}$ is the covariance of the measurement noise $W_t$ and the norm is with respect to the prior measure $\rho_0$. The posterior measure is given by $\rho(dx)\propto \exp(-\phi(x))\rho_0(dx)$. Upon linearization of $\nabla f(x)$, we may write a locally quadratic approximation of Eqn. \eqref{eqn:Phi} as:
\begin{equation}
    \phi_L(x)=\frac{1}{2}||\boldsymbol{Q}^{-1/2} \boldsymbol{H}x ||^2
\end{equation}
$\boldsymbol{H}$ is the Hessian of $f$. As shown in \cite{schillings2017analysis}, the continuous time limit of the recursive update of $x$ may be approximated as 
\begin{equation}\label{eqn:gradflow}
    \frac{d\boldsymbol{X}^{j}}{dt} = -\boldsymbol{C}(\boldsymbol{X}) \nabla_{_{\boldsymbol{X}^j}} \phi_L(\boldsymbol{X}^{j})
\end{equation}
where $\boldsymbol{C}(\boldsymbol{X}) = \frac{1}{p} \sum_{j=1}^{p}(\boldsymbol{X}^j - \overline{\boldsymbol{X}}) (\boldsymbol{X}^j - \overline{\boldsymbol{X}})^T$ is the ensemble covariance and $\boldsymbol{X}^j$ denotes the $j^{th}$ column vector in the ensemble matrix $\boldsymbol{X}$. The gradient structure as above clearly ensures convergence of the updates whose rate is dictated by the real positive parts of the eigenvalues of the positive semi-definite matrix $\boldsymbol{C}(\boldsymbol{X})$. Also, note that $\phi_L(x)$ is a Lyapunov function for the dynamical flow defined by Eqn. \eqref{eqn:gradflow}. This is because $\frac{d \phi_L(\boldsymbol{X}^j)}{dt} = - \langle  \nabla_{_{\boldsymbol{X}^j}} \phi_L(\boldsymbol{X}^j), \nabla_{_{\boldsymbol{X}^j}} \phi_L(\boldsymbol{X}^j)\rangle_{\boldsymbol{C}(\boldsymbol{X})} \leq 0$.

In addition, by defining the particle-wise anomaly ${\boldsymbol{E}}_t^j={\boldsymbol{X}}_t^j-\overline{\boldsymbol{X}_t}$, where $\overline{\boldsymbol{X}_t}$ is the ensemble average at the pseudo-time $t$, it may be shown that a quadratic form of the locally propagated error, given by the matrix $\boldsymbol{\mathcal{E}}_{ij}=\langle \boldsymbol{HE}^i,\boldsymbol{HE}^j\rangle_{\boldsymbol{Q}}$, evolves according to $\frac{d}{dt}\boldsymbol{\mathcal{E}}=-\frac{2}{p}{\boldsymbol{\mathcal{E}}}^{2}$. Hence $\boldsymbol{\mathcal{E}}$ converges to $[0]_{p\times p}$ as $t \rightarrow \infty$. In other words, all the particles asymptotically collapse to the empirical mean.

Following the discretized form of the filtering dynamics as in Eqn. (\ref{e5}) and a supermartingale structure of the evolving $f(X_t^1)$, it should also be possible to invoke Doob's martingale convergence theorem \cite{klebaner2012introduction} to prove convergence of our scheme. Indeed, a closer look at the approximations in the gain matrix reveals that the supermartingale structure of $f(X_t^1)$ (or an appropriate subsequence of $f(X_t^1)$), which is an essential element for convergence, is not interfered with. Details of such a proof will be taken up in a future article. 

\section{Results and Discussion}\label{s3}
This section is on a numerical demonstration of the feasibility and efficacy of FINDER in training wide and deep neural networks. These networks pose significant challenges \cite{Goodfellow-et-al-2016} due to presence of multiple local minima and vanishing or exploding gradients. This renders traditional gradient search mostly ineffective. In this context, physics informed neural network problems constitute a class of optimization problems with a challenging loss landscape due to presence of derivatives in the loss function. Besides, by comparing the solution via PINNs with the analytical solution if available, one may readily assess the optimizer performance. We solve a number of such problems among others, but with increased width and depth of network. For starters, we test the performance of FINDER on several IEEE benchmark functions (mostly in relatively high dimensions). This is followed by the problem of training wide and deep PINNs. Although we assume the loss function to be deterministic throughout Section \ref{s2}, we also demonstrate FINDER's ability to handle noisy losses. Towards this, We present a modified version of the algorithm and apply it to the MNIST classification problem.

All numerical experiments were conducted on a system equipped with 32GB RAM and 4GB NVIDIA T1000 GPU. The programs are available online on  \href{https://github.com/FINDER-optimizer/FINDER}{github}; they are written in Python using standard open-access libraries. 
%\st{We have used the following values of hyper-parameters: $p=5,\;\theta = 0.9,\; \gamma = 1,\; c_s = 0.1,\; c_\alpha = 0.01$ in all problems. $\zeta_1 = \zeta_2 = 0.0001$ unless stated otherwise.}
For all comparisons with the Adam optimizer \cite{kingma2014adam}, a constant learning rate of $10^{-3}$ is used and the default values of the hyper-parameters are:  $\beta_1=0.9$, $\beta_2=0.999, \epsilon = 10^{-8}$. Similarly, in all comparisons with L-BFGS, a strong Wolfe condition based on cubic interpolation line search scheme is used (as available in the pytorch library). For quick reference, the runtime of each optimizer is mentioned in the plot legends. A $^*$ mark over the indicated time means that optimality was not achieved in the stated interval.

\subsection{IEEE benchmark problems}
\begin{figure}
    \centering
\includegraphics[scale=0.4]{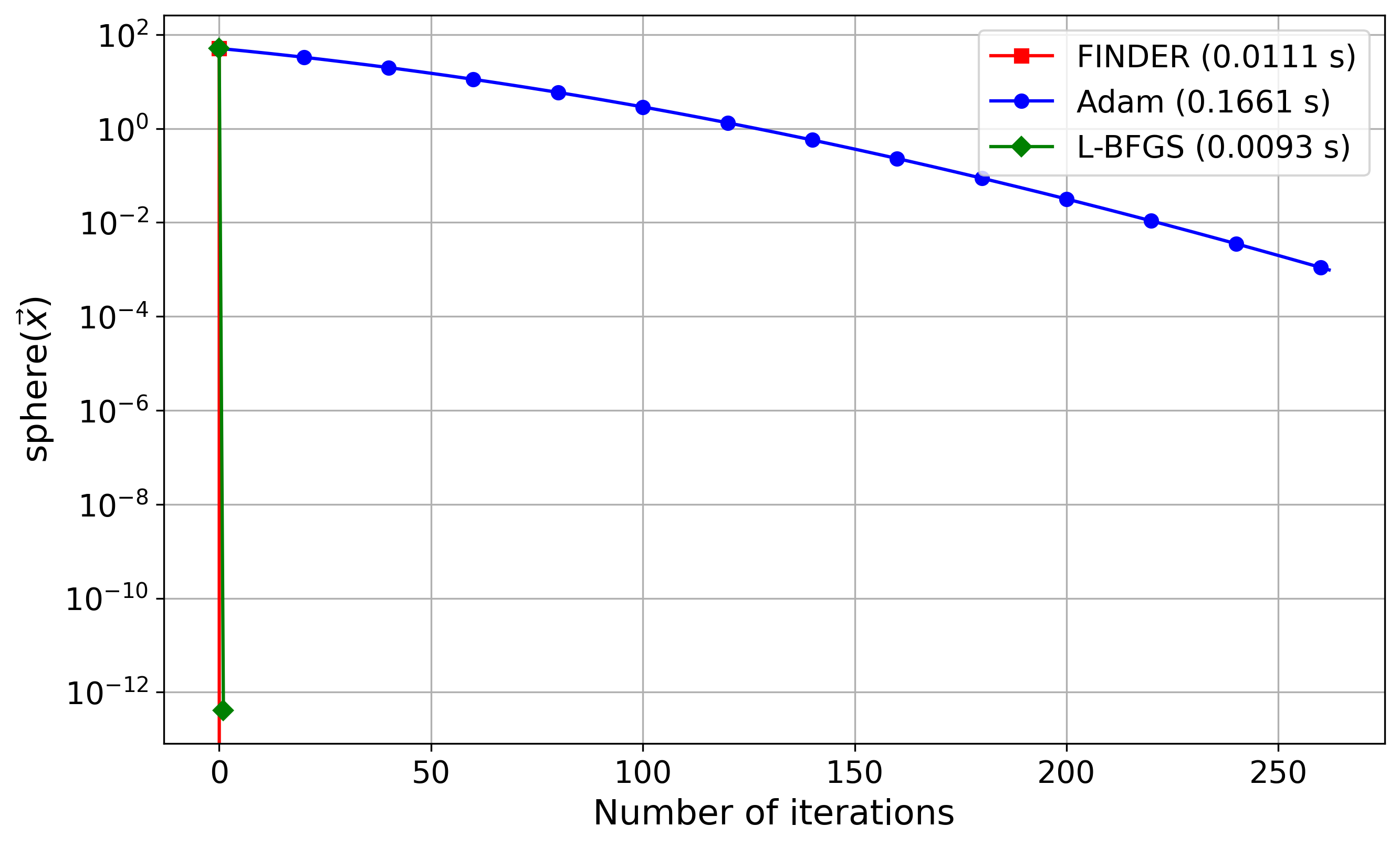}
    \caption{Minimization of 5000-dimensional sphere function}
    \label{fig:sphere2}
\end{figure}

\begin{figure}
    \centering
\includegraphics[scale=0.4]{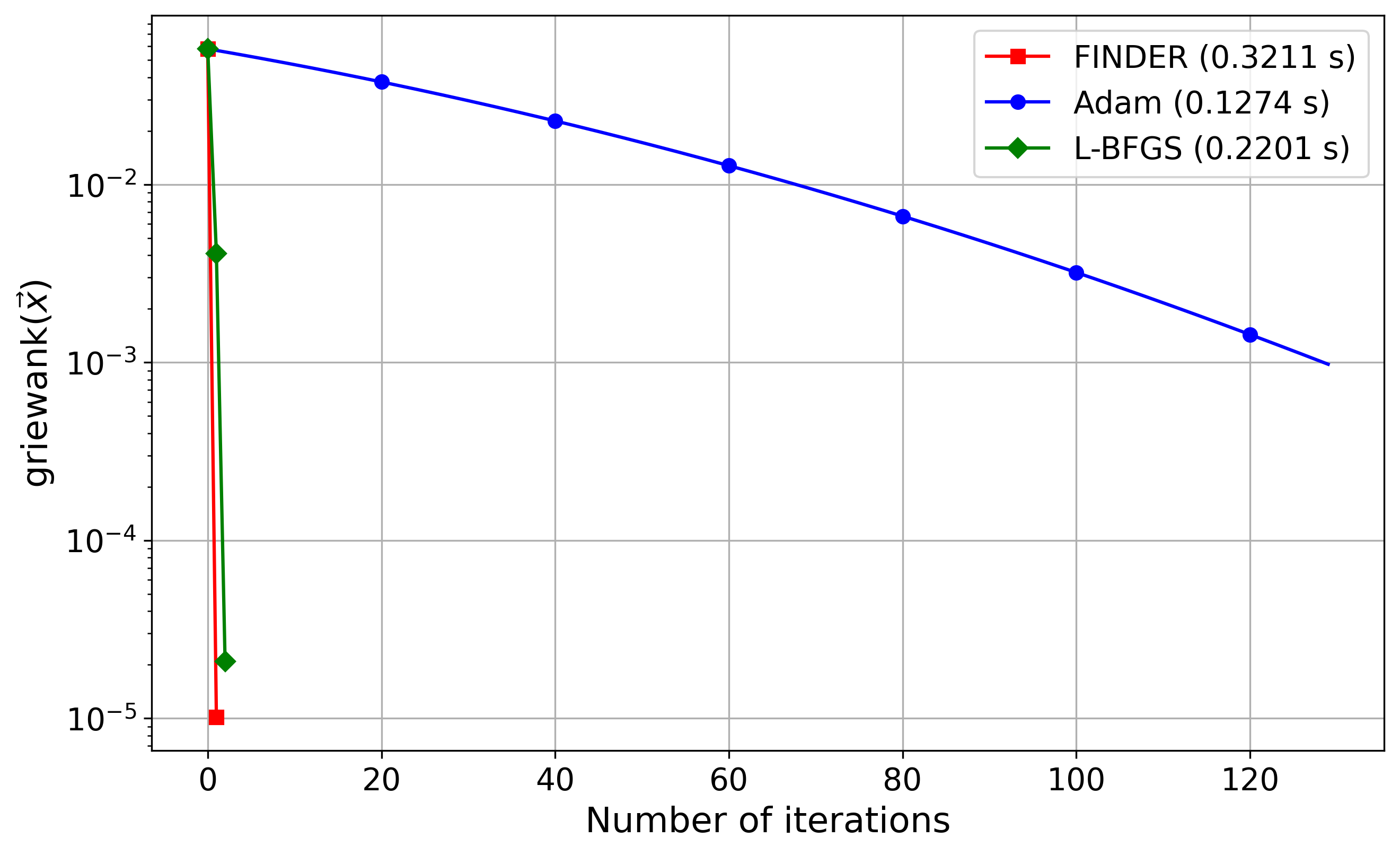}
    \caption{Minimization of the 5000-dimensional Griewank function}
    \label{fig:griewank}
\end{figure}

\begin{figure}
    \centering
\includegraphics[scale=0.4]{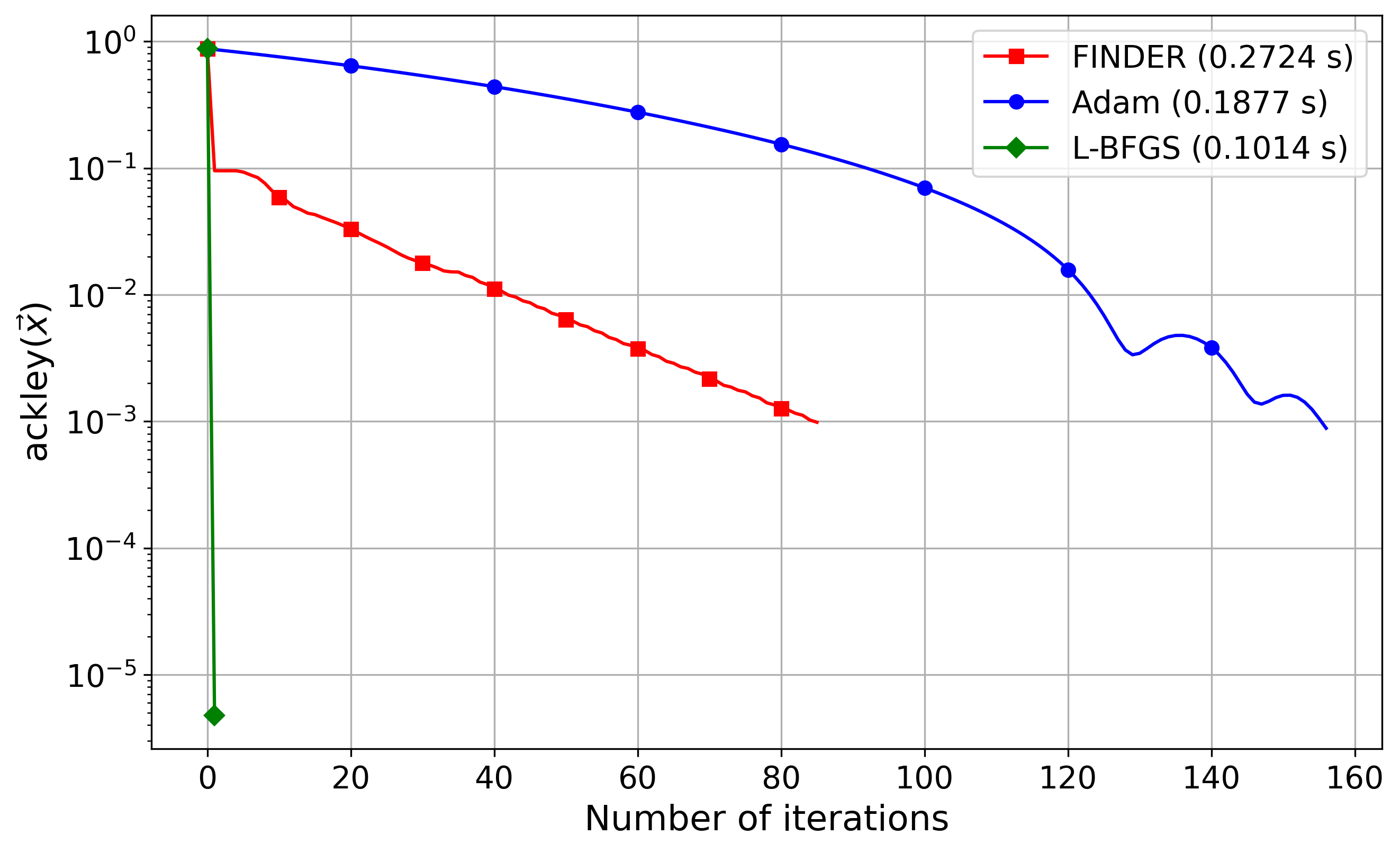}
    \caption{Minimization of 5000-dimensional Ackley function}
    \label{fig:ack2}
\end{figure}

\begin{figure}
    \centering
\includegraphics[scale=0.4]{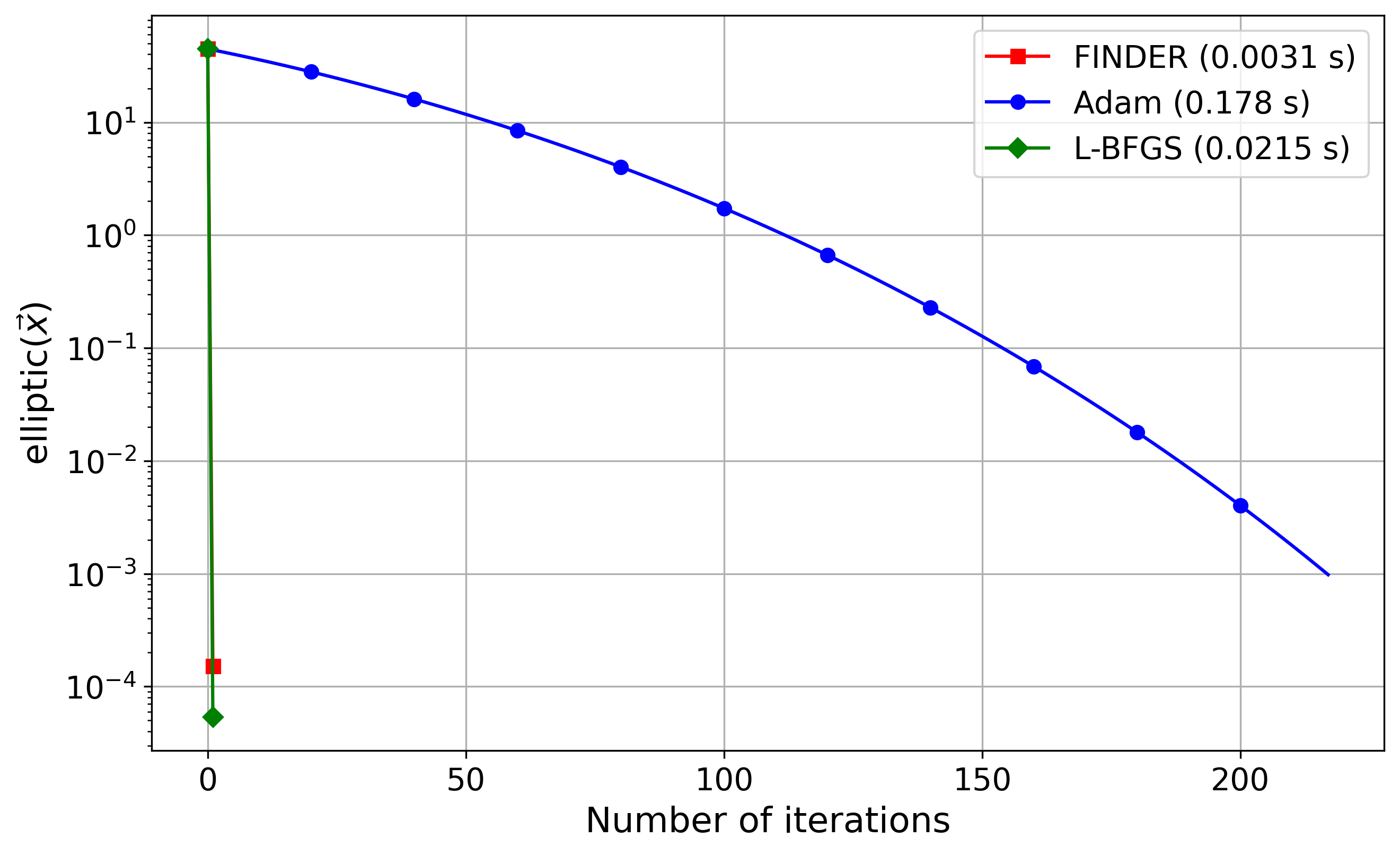}
    \caption{Minimization of 5000-dimensional high-Conditioned Elliptic function}
    \label{fig:hcef}
\end{figure}

\begin{figure}
    \centering
\includegraphics[scale=0.4]{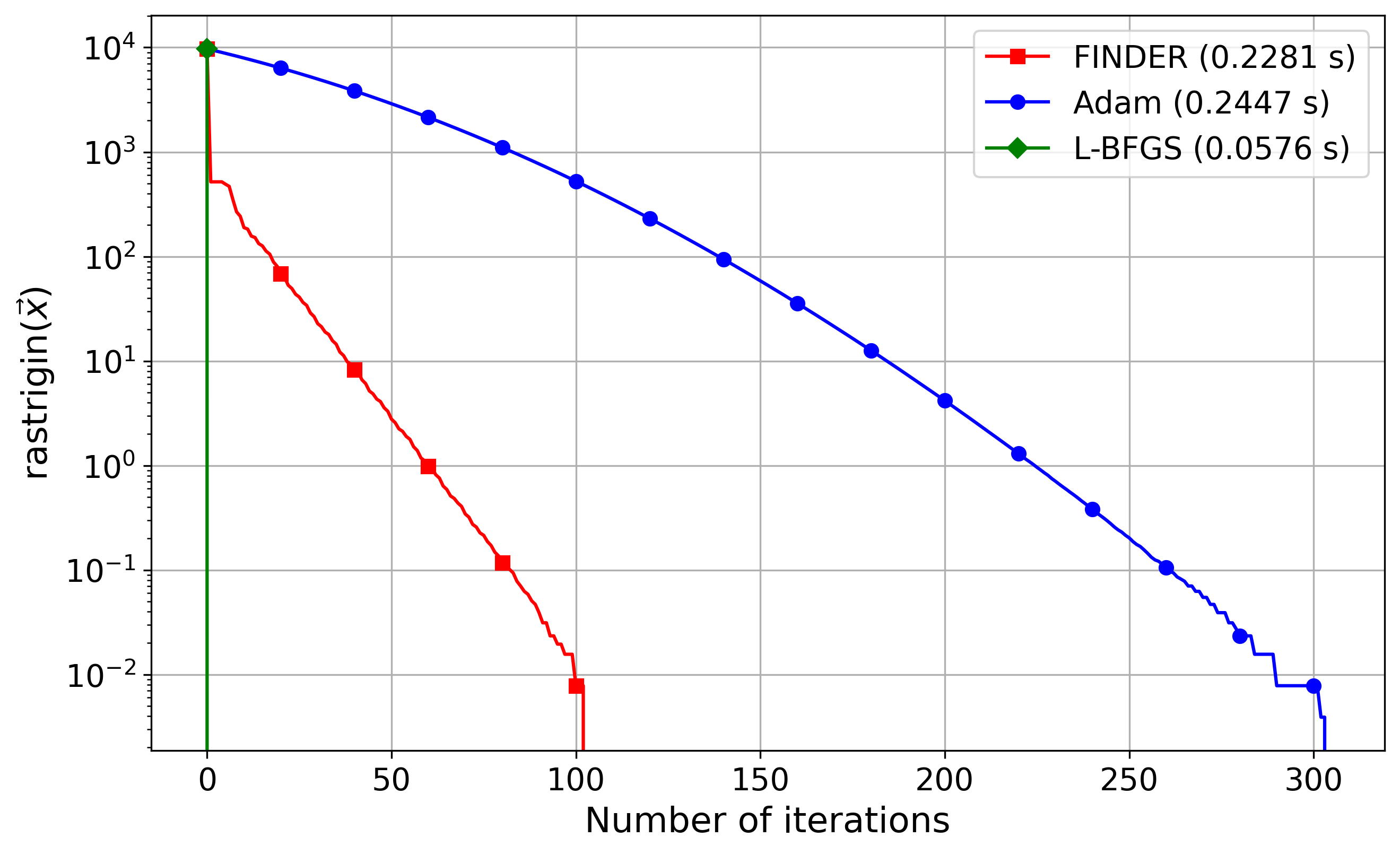}
    \caption{Minimization of 5000-dimensional Rastrigin function}
    \label{fig:rast2}
\end{figure}

\begin{figure}
    \centering
\includegraphics[scale=0.4]{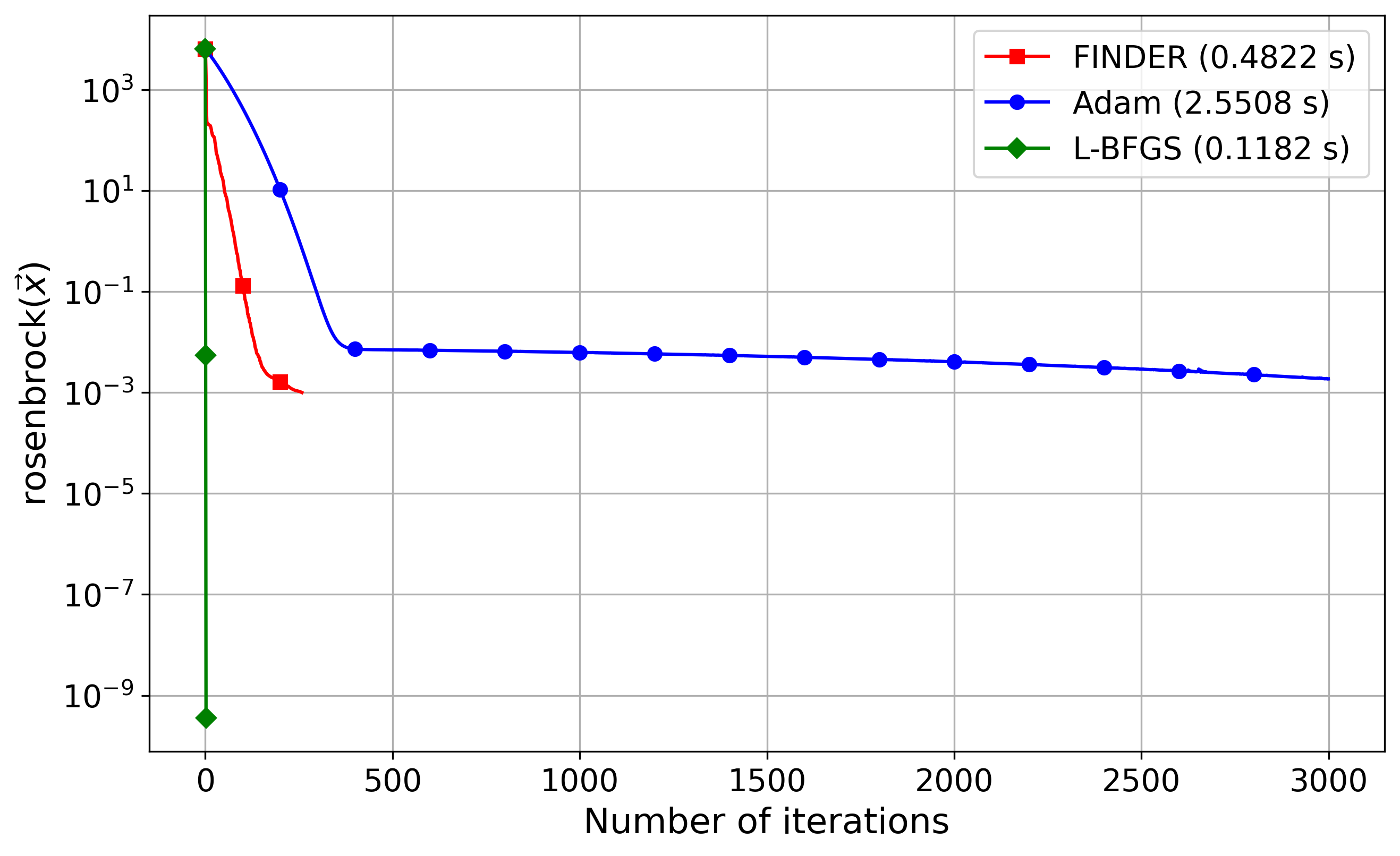}
    \caption{Minimization of 5000-dimensional Rosenbrock function}
    \label{fig:rosen2}
\end{figure}

The IEEE benchmark problems \cite{plevris2022collection} are widely recognized as a standard point of reference to test and compare different optimization algorithms. 
From the results on minimizing Ackley, Rosenbrock and Griewank functions, it is clear that FINDER works well even when the true Hessian has non-zero off-diagonal terms (See Figs.  \ref{fig:sphere2}-\ref{fig:rosen2}). Our attempts to use certain well-known stochastic search algorithms, e.g.  GA\cite{srinivas1994genetic}, PSO\cite{wang2018particle}, CMA-ES\cite{hansen2006cma}, for these benchmark functions have not produced converged solutions for higher problem dimensions, e.g. those exceeding $1000$.  


% The goal here is to correctly categorize each image from the CIFAR10 dataset based on the object captured in that image. We select a Convolutional Neural Network (CNN) as our model, comprising a single convolutional layer featuring 16 kernels and filters sized $5\times5$. This layer undergoes a max-pooling operation with $relu$ activation, followed by a feed-forward neural network with 240, 84, and 10 neurons; each hidden layer employs $\tanh$ activation and the output layer employs $softmax$ activation. 
% The loss function evolutions over 100 epochs are shown in Fig. \ref{fig:Cipha10} 
% \begin{figure}
%     \centering
% \includegraphics[scale=0.4]{Images/cifar10/cifar10.png}
%     \caption{Minimization of the loss function for CIFAR10 classification problem; $N = 775190$}
%     \label{fig:Cipha10}
% \end{figure}

\subsection{Training Physics Informed Neural Networks (PINNs)}
Motivated by smooth approximations of functions and their derivatives in neural networks, PINNs use physics information, such as squared residuals of the governing ordinary or partial differential equations (ODEs/PDEs) and initial/boundary conditions (ICs/BCs) to construct loss functions. Loss functions are then minimized to determine the network parameters. In all the examples shown here, the BCs/ICs have been imposed by modifying the network output itself.

\subsubsection{Burgers' Equation}
\begin{figure}
    \centering
\includegraphics[scale=0.4]{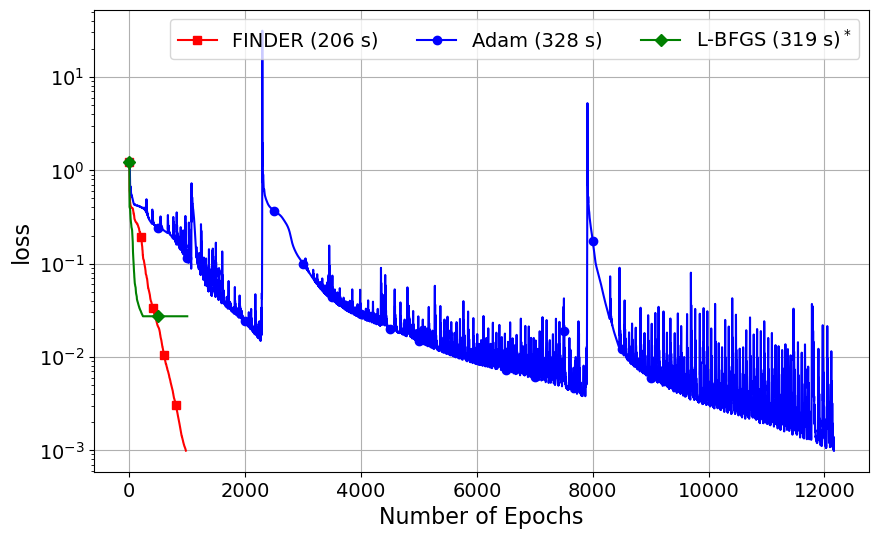}
    \caption{Minimization of the loss function for Burgers' equation; $N = 20601$}
    \label{fig:burger}
\end{figure}

\begin{figure}
    \centering
\includegraphics[scale=0.4]{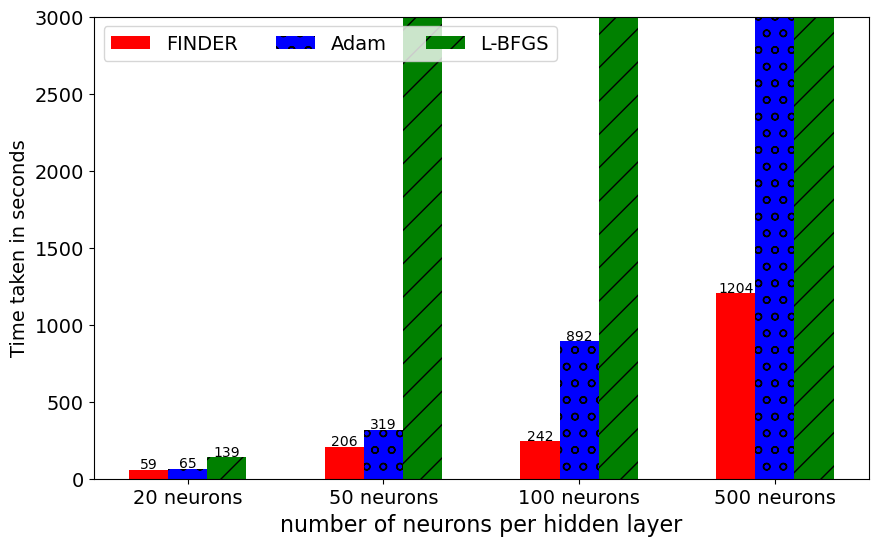}
    \caption{Effect of network width on optimizer performances for Burgers' problem}
    \label{fig: barplot}
\end{figure}

In the low-viscosity regime, the solution to this equation may lead to the formation of a shock, a phenomenon generally difficult to resolve using traditional numerical methods. We specifically consider the following problem \cite{raissi2017physics}:
\begin{equation}\label{burgers}
    \begin{aligned}
        u_{,t} +uu_{,x}&=\frac{0.01}{\pi}u_{,xx}\\ 
        u(0,x)&=-\sin(\pi x)\rightarrow\text{IC}\\
        u(t,-1)&=u(t,1)=0\rightarrow\text{BCs}\\
    \end{aligned}
\end{equation}
where $x \in [-1,1]$ and $t \in [0,1]$. Details of the PINN formulation of Eq. \eqref{burgers} is given in the Supplementary Material. 
% \begin{equation}
%     \begin{aligned}
%         u &\leftarrow\mathbf{DNN}(t,x; \Theta)\\
%         \hat{u} &= t(x^2 - 1)u - sin(\pi x)\\
%         loss &= \frac{1}{N_c}\sum_{i=1}^{N_c} (\hat{u}^{(i)}_{,t} +\hat{u}^{(i)}\hat{u}^{(i)}_{,x}-\frac{0.01}{\pi}\hat{u}^{(i)}_{,xx})^2\\
%         N_c &= 10000
%     \end{aligned}
% \end{equation}
% where $\Theta$ denotes trainable parameters of the DNN. 
The field variable $u$ is treated as the output of a DNN with two input neurons (each for $x$ and $t$) and 9 hidden layers, each with 50 neurons and $Tanh$ activation. $10000$ interior points are drawn from the domain using the uniform distribution $\mathcal{U}[-1,1]\times\mathcal{U}[0,1]$. Boundary and initial conditions are enforced by modifying the network output itself. The loss function is thus just the mean of squared PDE residuals evaluated at collocation points.
%This is a highly non-linear and non-convex problem due to presence of both first order and second order derivatives in the loss function. 
The model's non-linearity is compounded by the incorporation of BCs/ICs into the network output. Note that we 
have increased the number of neurons per layer from 20 to 50 vis-\'a-vis \cite{raissi2017physics}. While L-BFGS manages to reach the optimal minimum faster than FINDER (both time and iteration count-wise) for DNN with 20 neurons per layer, it gets trapped in a sub-optimal minimum when DNN with 50 neurons per layer is used. As evident from Fig. \ref{fig:burger}, FINDER proves more resilient to the challenges posed by wide DNNs, exhibiting quasi-Newtonian descent (without spiky evolutions seen under Adam), whilst consistently avoiding local minima traps.
We further highlight the strength of FINDER for wide networks by varying the number of neurons per layer while keeping the number of layers fixed.  We also compare the time taken by each optimizer to reach a tolerance limit in Fig. \ref{fig: barplot}.
Clearly, the increase in time taken to reach the tolerance limit is almost linear for FINDER, while it is not only quadratic for Adam but also fails at 500 neurons per layer.

\subsubsection{2D Elasticity}
In the mechanics of solid continua, the 2D plain strain elasticity problem describes the instantaneous deformation of a two-dimensional body under external forces, with no deformation in the third dimension. We attempt to solve this problem \cite{haghighat2021physics} using PINN. The governing PDEs, constitutive laws, and kinematic relations are as follows: 
\begin{equation}\label{elasticity}
    \begin{aligned}
        \text{Linear momentum balance: }&\sigma_{xx,x}+\sigma_{xy,y}+b_{x}=0\\        &\sigma_{yx,x}+\sigma_{yy,y}+b_{y}=0\\        
        &\sigma_{xy}=\sigma_{yx}\\
        \text{Constitutive laws: }&\sigma_{xx}=(\lambda+ 2\mu)\varepsilon_{xx}+\lambda\varepsilon_{yy}\\
        &\sigma_{yy}=(\lambda+ 2\mu)\varepsilon_{yy}+\lambda\varepsilon_{xx}\\
        &\sigma_{xy} = 2\mu\varepsilon_{xy}\\
        \text{Kinematic identities: }&\varepsilon_{xx}=u_{,x};\varepsilon_{yy}=v_{,y}\\
        &\varepsilon_{xy} = \frac{1}{2}\left(u_{,y}+v_{,x}\right)\\
        \text{Body force components: }&\\
        b_{x}=\lambda(4\pi^2\cos(2\pi x)\sin(\pi y)&-\pi \cos(\pi x)\mathcal{Q}y^3)\\
        +\mu(9\pi^2cos(2\pi x)sin(\pi y)&-\pi cos(\pi x)\mathcal{Q}y^3)\\
        b_{y}=\lambda(-3sin(\pi x)\mathcal{Q}y^2+2\pi&^2sin(2\pi x)cos(\pi y))\\
        +\mu(-6sin(\pi x)\mathcal{Q}y^2+2\pi&^2sin(2\pi x)cos(\pi y))\\
        +\mu\pi^2sin(\pi x)\mathcal{Q}y^4/4\;\;\;\;\;\;\;&
    \end{aligned}
\end{equation}
where $\sigma$ denotes the stress tensor, $b$ the body force vector, $u,v$ scalar displacements along $x,y$ directions and $\varepsilon$ the infinitesimal strain tensor. We also take $\lambda=1$, $\mu=0.5$ and $\mathcal{Q}=4$.  A schematic of the BCs is shown in Fig. \ref{fig:elas}. The detailed PINN formulation of Eqn \eqref{elasticity} is given in the Supplementary Material. 
% \begin{equation}\label{pinn_elasticity}
%     \begin{aligned}
%         u,v, \sigma_{xx}, &\sigma_{yy}, \sigma_{xy}\leftarrow\mathbf{DNNs}(x,y;\Theta)\\
%         \hat{u} &= y(1-y)u\\
%         \hat{v} &= x(1-x)yv\\
%         \hat{\sigma}_{xx} &= x(1-x)\sigma_{xx}\\
%         \hat{\sigma}_{yy} &= (1-y)\sigma_{yy} + (\lambda + 2\mu)Q sin(\pi x)\\
%         % loss &= \frac{1}{N_c}\sum_{i=1}^{N_c}(\hat{\sigma}^{(i)}_{xx,x} + \hat{\sigma}^{(i)}_{xy,y} + b^{(i)}_x)^2\\
%         % &+ (\hat{\sigma}^{(i)}_{yx,x} + \hat{\sigma}^{(i)}_{yy,y} + b^{(i)}_y)^2\\
%         % &+ (\hat{\sigma}^{(i)}_{xx} - (\lambda + 2\mu)\hat{u}^{(i)}_{,x} - \lambda\hat{v}^{(i)}_{,y})^2\\
%         % &+ (\hat{\sigma}^{(i)}_{yy} - (\lambda + 2\mu)\hat{v}^{(i)}_{,y} - \lambda\hat{u}^{(i)}_{,x})^2\\
%         % &+ (\sigma^{(i)}_{xy} - \mu(\hat{u}^{(i)}_{,y} + \hat{v}^{(i)}_{,x})^2\\
%     \end{aligned}
% \end{equation}
% where, $\Theta$ denotes trainable network parameters. 
The PINN that we employ has separate DNNs for $u, v, \sigma_{xx}, \sigma_{yy}$ and $\sigma_{xy}$. Each DNN has 2 input neurons and 6 hidden layers each with 100 neurons and $tanh$ activation. The BCs are imposed by modifying the DNN outputs themselves. Input $(x,y)$ to the model is through a set of boundary and interior points.
\begin{figure}
    \centering
\includegraphics[scale=0.3]{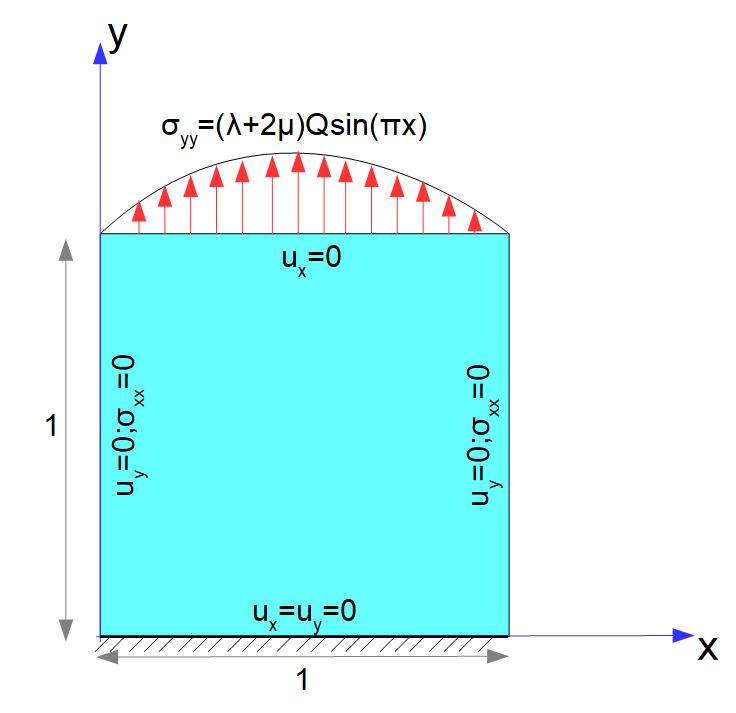}
    \caption{A schematic of domain, loading, and BCs for the 2D elasticity problem}\label{fig:elas}
\end{figure}
\begin{figure}
    \centering
\includegraphics[scale=0.4]{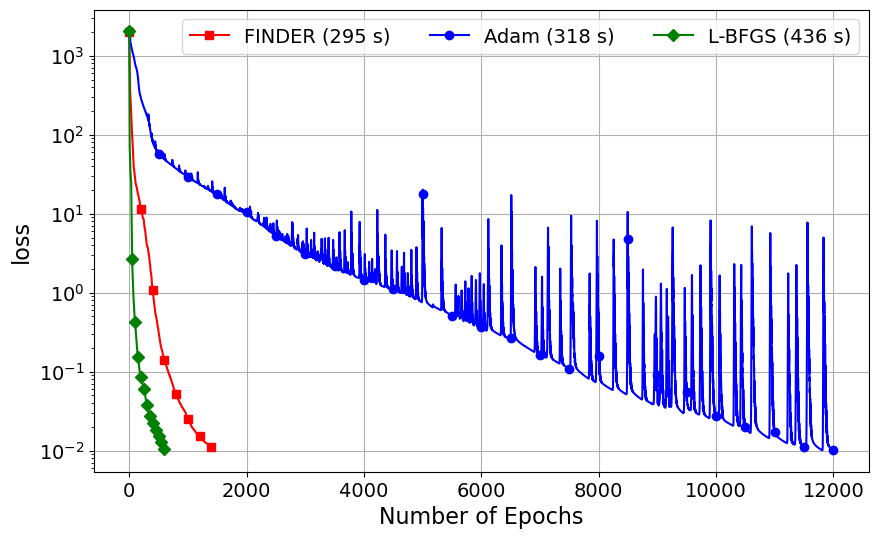}
    \caption{Minimization of the loss function for the 2D elasticity problem; $N=254505$. FINDER took less time but more epochs.}\label{fig:elasticity}
\end{figure}
The domain shown in Fig. \ref{fig:elas} is discretized as a $40\times 40$ grid resulting in 1600 points. The loss function is the summed residues of the governing equations evaluated at these points. Unlike the Burgers problem, we now have multiple networks and their outputs and derivatives integrated into a common loss function. We have increased the number of neurons per layer from 50 to 100 as compared to the original work \cite{haghighat2021physics}. Although L-BFGS converges to the minimum in fewer epochs, FINDER demonstrates faster training times compared to Adam and L-BFGS. Evolutions of the loss function in the three optimizers are plotted in Fig. \ref{fig:elasticity}. It is evident that FINDER achieves a quasi-Newtonian descent with no spiky evolution as seen with Adam.

\begin{figure}
    \centering
\includegraphics[scale=0.4]{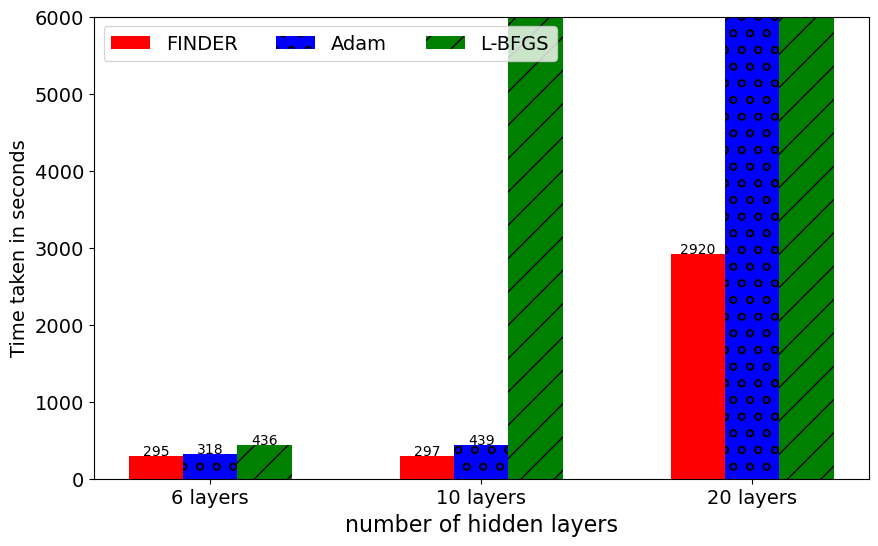}
    \caption{Effect of network depth on optimizer performance for 2D elasticity problem.}\label{fig:elasticity_bar}
\end{figure}

In Fig. \ref{fig:elasticity_bar}, we also compare the performance of optimizers for varying network depth and a fixed width (100 neurons per hidden layer).

\subsubsection{Strain Gradient Plasticity}
\begin{figure}
    \centering
\includegraphics[scale=0.4]{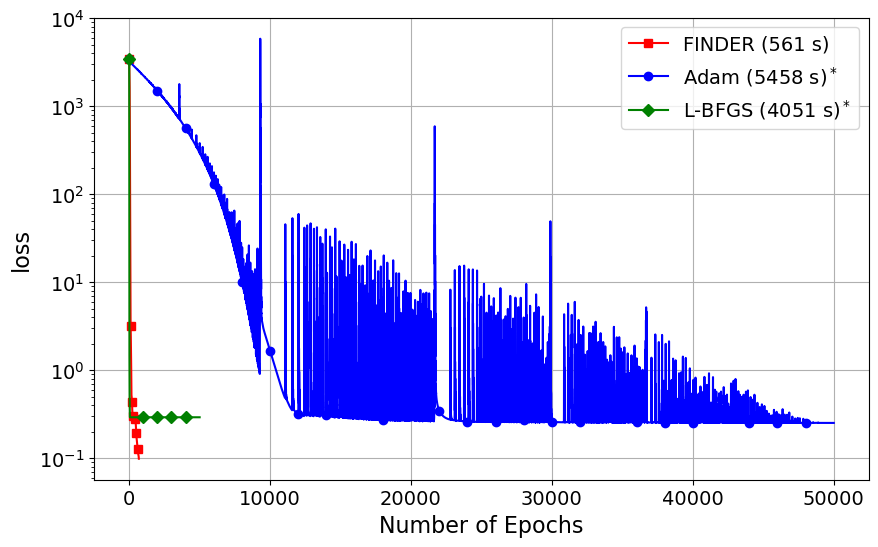}
    \caption{Minimization of the loss function for the strain gradient plasticity problem; $N=182602$}
    \label{fig:plasticity}
\end{figure}

Strain gradient plasticity is a mathematical model that acknowledges the importance of plastic strain gradients, thereby accounting for intrinsic (microscopic) length scales, in modeling plastic deformation in solids, e.g. metals. Classical plasticity theories, without such length scales, exhibit spurious mesh-dependence of solutions (viz. shear bands) when implemented numerically. 
% \st{These include strain softening/hardening, where some materials, upon stretching, lose strength while others gain; size effect, where smaller material samples exhibit higher strength than the bigger ones; inhomogeneous plastic flow, where materials do not stretch evenly; cases of very large plastic deformation etc. These features are far too macroscopically manifest to be efficiently simulated using molecular dynamics. This creates a gap between computational feasibility and theoretical modelling. With its multi-scale features, the strain gradient plasticity model }\cite{gurtin2005theory}\st{ has been an attempt to bridge the gap.} 
The multi-scale and extremely stiff nature of the model is captured by the governing macroscopic and microscopic balance laws, which takes the form of a power law; see Supplementary Material for details.
% Specifically, the microscopic balance is designed to evolve the plastic strain rate $\dot{\gamma}^p$ with $\tau^p$ denoting the stress conjugate. This balance law also involves a term $\nabla\dot{\gamma}^p$ -- the gradient of plastic strain rate, and the $k^p$ is the associated stress conjugate. Attempts at directly solving these partial differential equations (balance laws) via the finite element method have generally been unsuccessful. 
A PINN-based approach has recently shown promise \cite{tyagi2024physics} with the use of a learning rate scheduler to reach the minimum. We now demonstrate how the notion of scheduling is implemented in FINDER.
As before, we employ all three optimizers to solve the stiff displacement boundary value problem (see Appendix A of \cite{lele2008class}) via PINN and compare their relative performance. The loss function has multiple local minima in proximity, leading to possible variations in results with different initializations of network parameters.
The governing equations and ICs/BCs are provided in the Supplementary Material.
% \begin{equation}
%     \begin{aligned}
%         \text{Macroforce balance: }&\tau_{,y} = 0\\
%         \text{Microforce balance: }&\tau = \tau^p - k^p_{,y}\\
%         \text{Constitutive law: }&\tau = \mu(u_{,y} - \gamma^p)\\ \text{Microstress: }&\tau^p = S_0\left(\frac{d^p}{d_0}\right)^m\frac{\dot{\gamma}^p}{d^p}\\
%         \text{Gradient microstress: }&k^p = SL^2\gamma^p_{,y} + S_0 l^2\left(\frac{d^p}{d_0}\right)^m\frac{\dot{\gamma}^p_{,y}}{d^p}\\
%         \text{Resistance to plastic flow: }&\dot{S} = H(S)d^p\text{ ; }S(y,0) = S_0\\
%         \text{Effective flow rate: }&d^p = \sqrt{\mid\dot{\gamma}^p\mid^2 + l^2\mid\dot{\gamma}^p_{,y}\mid^2}\\
%         \text{Imposed shear strain: }&E(t) = \frac{u^{\text{\cross[0.4pt]}}(t)}{h}\\
%         \text{Displacement BCs: }&u(0,t) = 0;\; u(h,t) = u^{\text{\cross[0.4pt]}}(t)\\
%         \text{Plastic strain rate BCs: }&\dot{\gamma}^p(0,t) = \dot{\gamma}^p(h,t) = 0\\
%         \text{Displacement ICs: }&u(y,0) = 0\\
%         \text{Plastic strain ICs: }&\gamma^p(y,0) = 0\\
%     \end{aligned}
% \end{equation}
% where $\tau$ denotes shear stress and $y\in [0,h]$. The problem involves a long (height $h$ along $y$), thin body (infinite in $x$ and $z$) undergoing plane-strain shearing along $x$, so that the displacement $u:=u(y,t)$. We use the initial yield strength $S_0 = 100$ MPa, shear modulus $\mu = 100$ GPa, hardening/ softening function $H = 0$, reference flow rate $d_0 = 0.1$, rate sensitivity parameter $m = 0.02$, energetic length scale $L = 10$, dissipative length scale $l = 0$ and $h = 10$.
In contrast to the original work \cite{tyagi2024physics} which uses just three hidden layers with 32 neurons each, we employ separate DNNs for $u$ and
$\gamma^p$, each with 2 input neurons and 10 hidden layers with 100 neurons per layer. $tanh$ activation is used in the hidden layers. The BCs and ICs are `hard-coded' by modifying the network output itself. We follow the same discretization of domain and scalings to inputs $t,y$ and outputs $u,\gamma^p$ as suggested in \cite{tyagi2024physics}. 
% \st{before constructing the loss function, which is the mean squared error - same as before?? residue of the given equations evaluated at 9702 collocation points, uniformly distributed in the domain.} 
The strategy for dealing with loss plateau is to use a learning rate scheduling scheme for each epoch. This is readily available for Adam under Pytorch. We show the implementation of a similar scheduling scheme for FINDER in which $\zeta_1,\zeta_2$ are initialized at 0.1 and linearly reduced to 0.00001 over 1000 epochs. Fig. \ref{fig:plasticity} shows evolutions of the loss function via FINDER, L-BFGS, and Adam. One observes that FINDER readily surpasses the loss plateau to attain quasi-Newtonian descent in about $(1/10)^{th}$ of the time taken by Adam and L-BFGS which remain trapped even after 50000 epochs. This is despite the use of a learning rate scheduler for Adam as suggested in \cite{tyagi2024physics}.

\subsection{A classification problem using neural networks}
Classifications are supervised learning problems based on data. During their training, data are often passed to the network in mini-batches, making the loss function noisy. The smaller the batch size, the higher the variance in gradient estimation. While L-BFGS fails to optimize small batch sizes, Adam manages to perform very well in such problems. A close look at the FINDER algorithm suggests that $\theta$, the convolutional variable in Eqn. \eqref{increment} accumulates the effect of past gradients in the update. If we switch off this variable and put $\theta = 0$, then only current gradients are used and accumulated variance in gradients is reduced. Another source of variance is $B$. Noisy gradients lead to noisier gain. Therefore, in order to balance out the effect of $B$ as a component-wise step size multiplier on the gradient vector, we use smaller values of $\gamma$. The third source of variance in FINDER is in its sampling strategy, which is dictated by $\zeta_1, \zeta_2$. Higher values of $\zeta_1, \zeta_2$, result in high oscillations in minimizing noisy loss functions. This is because the fittest particle for the loss function of one batch could be unfit for that of another batch.  Based on our parametric study of FINDER on the MNIST classification problem, we recommend using $\zeta_1=\zeta_2 = 10^{-6}; \theta = 0$ for all noisy objectives. $\gamma = 0.1$ for batch size of 100 and $\gamma= 0.5$ for batch size of 500 are found to achieve neraly fluctuation-free descent. 

% Classification problems are of immense significance in machine learning and artificial intelligence. By accurately recognizing labels (such as digits, animals, cars etc.) from images, they demonstrate a model's ability to understand complex patterns and make decisions based on the input data. In this context, datasets like MNIST \cite{lecun2010mnist} \& CIFAR10 \cite{Krizhevsky09learningmultiple} provide a standardized platform for comparing the efficacy of different algorithms. In our implementation, a single batch of training data has been fed to the networks as input. Gradient computation is presently done through the automatic differentiation toolkit `Autograd' of PyTorch \cite{paszke2017automatic}. We have trained for 100 epochs without prescribing any $\varepsilon_{tol}$ in these problems.

\subsubsection{MNIST classification problem}
\begin{figure}
    \centering
\includegraphics[scale=0.4]{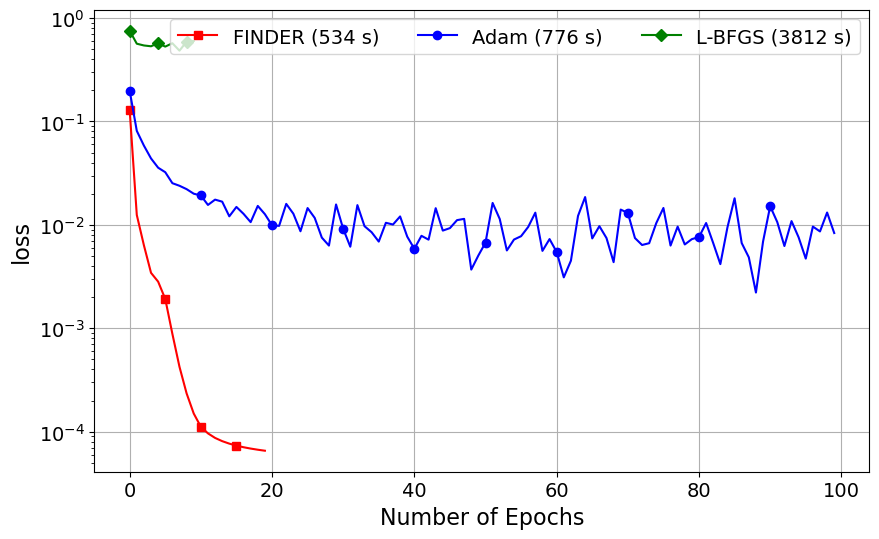}
    \caption{Minimization of the loss function for MNIST problem;$N=1796010$}.
    \label{fig: MNIST2}
\end{figure}

\begin{figure}
    \centering
\includegraphics[scale=0.4]{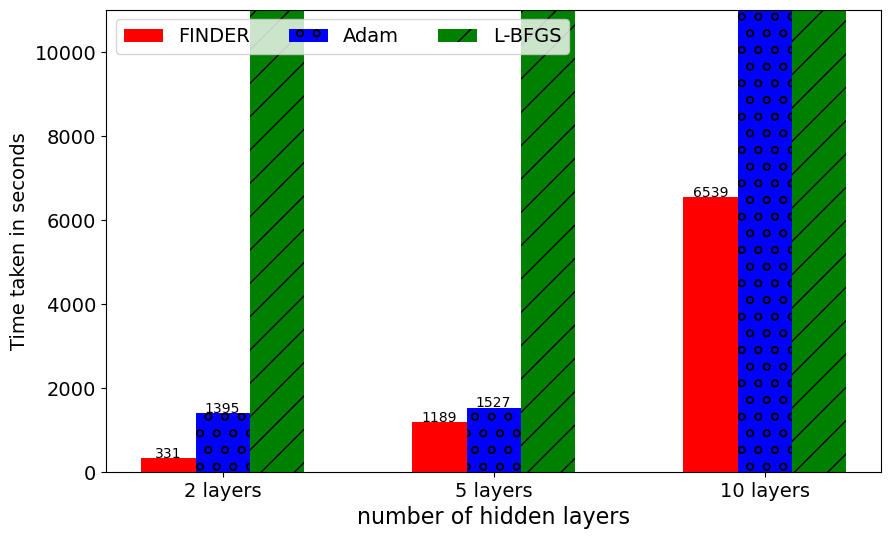}
    \caption{Effect of network depth on optimizer performance for the MNIST classification problem.}\label{fig:mnist_bar}
\end{figure}

Our goal is to train a DNN to correctly identify the digit in each image from the MNIST dataset \cite{lecun2010mnist}. We choose the same network architecture and batch size as used in Kingma \textit{et al.} \cite{kingma2014adam} which involves two hidden layers of 1000 neurons each with $ReLU$ activation on the hidden layers and each batch with 128 training images. For the hyperparameters in FINDER, we use $\theta = 0, \gamma = 0.1, \zeta_1=\zeta_2 = 10^{-6}$. Fig. \ref{fig: MNIST2} shows a comparison with the Adam optimizer. FINDER achieves the speed of quasi-Newton descent and loss value of an order less in less time and less number of epochs vis-\'a-vis Adam. Adam has an oscillating descent centred around a loss value of order $10^{-2}$ and the same order of loss is obtained by FINDER in $(1/8)^{th}$ of the time taken by Adam. Accuracy on the training set is 100\% with FINDER after 20 epochs and 99.92\% with Adam after 100 epochs. Accuracy on the test set is 98.73\% with FINDER and 98.22\% with Adam.
In Fig. \ref{fig:mnist_bar}, we also compare the performance of optimizers for variable network depth and a fixed width, i.e. 1000 neurons per hidden layer.
\subsection*{Discussion}
As such, FINDER's computing time per epoch is substantially higher than Adam's; for every epoch, the latter is about 10 times faster. This may be attributed to the fact that FINDER performs $2p$ forward passes and $p$ backward passes per epoch while Adam does one forward pass and one backward pass per epoch. Furthermore, the backtracking line search to determine the step size entails additional forward passes in every epoch. Thus, in its present form, FINDER is not very competitive for convolutional neural networks (CNNs). As CNNs deepen with more convolutional layers, the forward pass overhead becomes overwhelming, causing FINDER to experience significant slowdown.
We expect that a GPU-based explicit parallelization of the multi-particle simulation or an approach based on tensor networks \cite{jahromi2024variational,stoudenmire2016supervised, cichocki2014tensor} would render FINDER more computationally efficient in future.

\section{Conclusions}\label{s4}
Motivated in part by the fast convergence of quasi-Newton schemes over a locally convex part of an objective or loss function, we have used a measure-theoretic means to a stochastic mirroring of the inverse Hessian, leading to a novel optimization scheme -- FINDER. In the process, we have introduced certain simplifying approximations that enable the method to scale linearly with dimension. The probabilistic foundation of the method also implies a noise-driven exploratory character and thus FINDER could be thought of as a scheme astride the best of both worlds. The measure-theoretic feature, which enables a distributional -- and not pointwise -- interpretation of quantities such as the inverse Hessian and even the solution, derives from a stochastic filtering route, wherein the gradient of the loss function is driven to a zero-mean noise (and not to zero, as with most deterministic methods) via a change of measures. This distributional interpretation comes in handy while training physics-informed deep networks wherein the presence of higher order derivatives may render the loss function spiky and render a pointwise interpretation of the gradient less meaningful. This probably explains why Adam, which uses the classical notion of gradient, exhibits (unlike FINDER) spiky evolutions for PINN implementations. 

As with most other optimizers, FINDER has its share of limitations and issues, at least in its current form. While it shows a consistently faster local convergence, the noise-assisted avoidance of local traps in FINDER appears to be less effective than the exploitation of the so-called variance of the stochastic gradient in Adam. Borrowing such concepts from Adam or using Levy noises \cite{applebaum2009levy} with or without annealing to avoid local entrapment could therefore be explored. FINDER uses multi-particle simulation and hence high computational overhead. Innovative decomposition of the weight tensor, as in tensor networks \cite{jahromi2024variational,stoudenmire2016supervised, cichocki2014tensor}, could be a way out of this difficulty.
%Curiously enough, the gain matrix multiplying the vector-valued innovation (i.e. the error process or the gradient of the loss function being driven to a zero-mean martingale) to form the update term is then identifiable as the mimicked inverse Hessian. 
%Moreover, based on the structure of our process and observation dynamics that form the backbone of the filtering problem, we may take the covariance of the estimated states to be diagonally dominant. This in turn would justify the diagonal approximation of the gain matrix by just neglecting the low-rank cross-covariance component. 
These limitations notwithstanding, results presented in this work show the remarkable promise of the method, perhaps uniquely so for a Monte Carlo search, across a spectrum of continuous optimization problems.
%The authors invite the readership to explore this optimizer for any continuous optimization requirements. 

\section*{Supplementary Material}
A derivation of the gain matrix used in update equation \eqref{e5} is provided in the supplementary material along with the details of the PINN examples covered in this article.
Python codes including a plug-and-play version are on \href{https://github.com/FINDER-optimizer/FINDER}{github}.
% \bibliographystyle{IEEEtran}
% \bibliography{biblio}

\begin{thebibliography}{10}
\providecommand{\url}[1]{#1}
\csname url@samestyle\endcsname
\providecommand{\newblock}{\relax}
\providecommand{\bibinfo}[2]{#2}
\providecommand{\BIBentrySTDinterwordspacing}{\spaceskip=0pt\relax}
\providecommand{\BIBentryALTinterwordstretchfactor}{4}
\providecommand{\BIBentryALTinterwordspacing}{\spaceskip=\fontdimen2\font plus
\BIBentryALTinterwordstretchfactor\fontdimen3\font minus \fontdimen4\font\relax}
\providecommand{\BIBforeignlanguage}[2]{{%
\expandafter\ifx\csname l@#1\endcsname\relax
\typeout{** WARNING: IEEEtran.bst: No hyphenation pattern has been}%
\typeout{** loaded for the language `#1'. Using the pattern for}%
\typeout{** the default language instead.}%
\else
\language=\csname l@#1\endcsname
\fi
#2}}
\providecommand{\BIBdecl}{\relax}
\BIBdecl

\bibitem{luenberger1984linear}
D.~G. Luenberger, Y.~Ye \emph{et~al.}, \emph{{Linear and nonlinear programming}}.\hskip 1em plus 0.5em minus 0.4em\relax Springer, 1984, vol.~2.

\bibitem{boyd2004convex}
S.~P. Boyd and L.~Vandenberghe, \emph{{Convex optimization}}.\hskip 1em plus 0.5em minus 0.4em\relax Cambridge University Press, 2004.

\bibitem{ruder2017overview}
S.~Ruder, ``{An overview of gradient descent optimization algorithms},'' 2017.

\bibitem{afshar2010size}
M.~Afshar and A.~Faramarzi, ``{Size optimization of truss structures by cellular automata},'' \emph{Journal of Computer Science and Engineering}, vol.~3, no.~1, pp. 1--9, 2010.

\bibitem{faramarzi2014novel}
A.~Faramarzi and M.~Afshar, ``{A novel hybrid cellular automata--linear programming approach for the optimal sizing of planar truss structures},'' \emph{Civil Engineering and Environmental Systems}, vol.~31, no.~3, pp. 209--228, 2014.

\bibitem{bertsekas1982projected}
D.~P. Bertsekas, ``{Projected {Newton} methods for optimization problems with simple constraints},'' \emph{SIAM Journal on control and Optimization}, vol.~20, no.~2, pp. 221--246, 1982.

\bibitem{ford1994multi}
J.~Ford and I.~Moghrabi, ``{Multi-step quasi-{Newton} methods for optimization},'' \emph{Journal of Computational and Applied Mathematics}, vol.~50, no. 1-3, pp. 305--323, 1994.

\bibitem{thapa1983optimization}
M.~N. Thapa, ``{Optimization of unconstrained functions with sparse {Hessian} matrices—Quasi-{Newton} methods},'' \emph{Mathematical Programming}, vol.~25, no.~2, pp. 158--182, 1983.

\bibitem{izmailov2014newton}
A.~F. Izmailov and M.~V. Solodov, \emph{{{Newton}-type methods for optimization and variational problems}}.\hskip 1em plus 0.5em minus 0.4em\relax Springer, 2014, vol.~3.

\bibitem{polyak2007newton}
B.~T. Polyak, ``{{Newton}’s method and its use in optimization},'' \emph{European Journal of Operational Research}, vol. 181, no.~3, pp. 1086--1096, 2007.

\bibitem{raissi2017physics}
M.~Raissi, P.~Perdikaris, and G.~E. Karniadakis, ``{Physics informed deep learning (part i): Data-driven solutions of nonlinear partial differential equations},'' \emph{arXiv preprint arXiv:1711.10561}, 2017.

\bibitem{haghighat2021physics}
E.~Haghighat, M.~Raissi, A.~Moure, H.~Gomez, and R.~Juanes, ``{A physics-informed deep learning framework for inversion and surrogate modeling in solid mechanics},'' \emph{Computer Methods in Applied Mechanics and Engineering}, vol. 379, p. 113741, 2021.

\bibitem{robbins1951stochastic}
H.~Robbins and S.~Monro, ``{A stochastic approximation method},'' \emph{The annals of mathematical statistics}, pp. 400--407, 1951.

\bibitem{srinivas1994genetic}
M.~Srinivas and L.~M. Patnaik, ``{Genetic algorithms: A survey},'' \emph{computer}, vol.~27, no.~6, pp. 17--26, 1994.

\bibitem{wang2018particle}
D.~Wang, D.~Tan, and L.~Liu, ``{Particle swarm optimization algorithm: an overview},'' \emph{Soft computing}, vol.~22, no.~2, pp. 387--408, 2018.

\bibitem{hansen2006cma}
N.~Hansen, ``{The CMA evolution strategy: a comparing review},'' \emph{Towards a new evolutionary computation: Advances in the estimation of distribution algorithms}, pp. 75--102, 2006.

\bibitem{rutenbar1989simulated}
R.~A. Rutenbar, ``{Simulated annealing algorithms: An overview},'' \emph{IEEE Circuits and Devices magazine}, vol.~5, no.~1, pp. 19--26, 1989.

\bibitem{storn1997differential}
R.~Storn and K.~Price, ``{Differential evolution--a simple and efficient heuristic for global optimization over continuous spaces},'' \emph{Journal of global optimization}, vol.~11, pp. 341--359, 1997.

\bibitem{kingma2014adam}
D.~P. Kingma and J.~Ba, ``{Adam: A method for stochastic optimization},'' \emph{arXiv preprint arXiv:1412.6980}, 2014.

\bibitem{nocedal1980updating}
J.~Nocedal, ``Updating quasi-newton matrices with limited storage,'' \emph{Mathematics of computation}, vol.~35, no. 151, pp. 773--782, 1980.

\bibitem{saxena2022microstructure}
M.~Saxena, S.~Sarkar, and D.~Roy, ``{A microstructure-sensitive and derivative-free continuum model for composite materials: applications to concrete},'' \emph{International Journal of Solids and Structures}, p. 112051, 2022.

\bibitem{oksendal2013stochastic}
B.~Oksendal, \emph{{Stochastic differential equations: an introduction with applications}}.\hskip 1em plus 0.5em minus 0.4em\relax Springer Science \& Business Media, 2013.

\bibitem{van2007stochastic}
R.~Van~Handel, ``Stochastic calculus, filtering, and stochastic control,'' \emph{Course notes., URL http://www. princeton. edu/rvan/acm217/ACM217. pdf}, vol.~14, 2007.

\bibitem{kushner1964differential}
H.~J. Kushner, ``{On the differential equations satisfied by conditional probablitity densities of Markov processes, with applications},'' \emph{Journal of the Society for Industrial and Applied Mathematics, Series A: Control}, vol.~2, no.~1, pp. 106--119, 1964.

\bibitem{stratonovich1968conditional}
R.~L. Stratonovich, R.~N. McDonough, N.~B. McDonough \emph{et~al.}, ``{Conditional Markov processes and their application to the theory of optimal control},'' \emph{IEEE Transactions on Automatic Control 13(1):137-138}, 1968.

\bibitem{sarkar2014kushner}
S.~Sarkar, S.~R. Chowdhury, M.~Venugopal, R.~M. Vasu, and D.~Roy, ``{A {Kushner--Stratonovich Monte Carlo} filter applied to nonlinear dynamical system identification},'' \emph{Physica D: Nonlinear Phenomena}, vol. 270, pp. 46--59, 2014.

\bibitem{saxena2024}
M.~Saxena, S.~Sarkar, and D.~Roy, ``{Inverse Hessian by stochastic projection and application to system identification in nonlinear mechanics of solids},'' \emph{International Journal of Non-Linear Mechanics}, p. 104762, 2024.

\bibitem{roy2017stochastic}
D.~Roy and G.~V. Rao, \emph{{Stochastic dynamics, filtering and optimization}}.\hskip 1em plus 0.5em minus 0.4em\relax Cambridge University Press, 2017.

\bibitem{bertsekas2021data}
D.~Bertsekas and R.~Gallager, \emph{Data networks}.\hskip 1em plus 0.5em minus 0.4em\relax Athena Scientific, 2021.

\bibitem{milstein2002monte}
G.~Milstein and J.~Schoenmakers, ``{{Monte Carlo} construction of hedging strategies against multi-asset {European} claims},'' \emph{Stochastics and Stochastic Reports}, vol.~73, no. 1-2, pp. 125--157, 2002.

\bibitem{yao2020adahessian}
Z.~Yao, A.~Gholami, S.~Shen, K.~Keutzer, and M.~W. Mahoney, ``{ADAHESSIAN: An Adaptive Second Order Optimizer for Machine Learning},'' \emph{AAAI (Accepted)}, 2021.

\bibitem{Nocedal2006}
\BIBentryALTinterwordspacing
J.~Nocedal and S.~J. Wright, \emph{{Line Search Methods}}.\hskip 1em plus 0.5em minus 0.4em\relax New York, NY: Springer New York, 2006, pp. 30--65. [Online]. Available: \url{https://doi.org/10.1007/978-0-387-40065-5_3}
\BIBentrySTDinterwordspacing

\bibitem{milstein2013numerical}
G.~N. Milstein, \emph{{Numerical integration of stochastic differential equations}}.\hskip 1em plus 0.5em minus 0.4em\relax Springer Science \& Business Media, 2013, vol. 313.

\bibitem{evensen2003ensemble}
G.~Evensen, ``{The ensemble Kalman filter: Theoretical formulation and practical implementation},'' \emph{Ocean dynamics}, vol.~53, pp. 343--367, 2003.

\bibitem{schillings2017analysis}
C.~Schillings and A.~M. Stuart, ``{Analysis of the ensemble Kalman filter for inverse problems},'' \emph{SIAM Journal on Numerical Analysis}, vol.~55, no.~3, pp. 1264--1290, 2017.

\bibitem{klebaner2012introduction}
F.~C. Klebaner, \emph{{Introduction to stochastic calculus with applications}}.\hskip 1em plus 0.5em minus 0.4em\relax World Scientific Publishing Company, 2012.

\bibitem{Goodfellow-et-al-2016}
I.~Goodfellow, Y.~Bengio, and A.~Courville, \emph{Deep Learning}.\hskip 1em plus 0.5em minus 0.4em\relax MIT Press, 2016, \url{http://www.deeplearningbook.org}.

\bibitem{plevris2022collection}
V.~Plevris and G.~Solorzano, ``{A collection of 30 multidimensional functions for global optimization benchmarking},'' \emph{Data}, vol.~7, no.~4, p.~46, 2022.

\bibitem{tyagi2024physics}
A.~Tyagi, U.~Suman, M.~Mamajiwala, and D.~Roy, ``{Physics Informed Deep Learning for Strain Gradient Continuum Plasticity},'' \emph{arXiv preprint arXiv:2408.06657}, 2024.

\bibitem{lele2008class}
S.~P. Lele, ``{On a class of strain gradient plasticity theories: formulation and numerical implementation},'' Ph.D. dissertation, Massachusetts Institute of Technology, 2008.

\bibitem{lecun2010mnist}
Y.~LeCun, C.~Cortes, and C.~Burges, ``{MNIST} handwritten digit database,'' \emph{ATT Labs [Online]. Available: http://yann.lecun.com/exdb/mnist}, vol.~2, 2010.

\bibitem{jahromi2024variational}
S.~S. Jahromi and R.~Or{\'u}s, ``Variational tensor neural networks for deep learning,'' \emph{Scientific Reports}, vol.~14, no.~1, p. 19017, 2024.

\bibitem{stoudenmire2016supervised}
E.~Stoudenmire and D.~J. Schwab, ``Supervised learning with tensor networks,'' \emph{Advances in neural information processing systems}, vol.~29, 2016.

\bibitem{cichocki2014tensor}
A.~Cichocki, ``Tensor networks for big data analytics and large-scale optimization problems,'' \emph{arXiv preprint arXiv:1407.3124}, 2014.

\bibitem{applebaum2009levy}
D.~Applebaum, \emph{L{\'e}vy processes and stochastic calculus}.\hskip 1em plus 0.5em minus 0.4em\relax Cambridge University Press, 2009.

\end{thebibliography}

\end{document}

\date{}
“This work has been submitted to the IEEE for possible publication. Copyright may be transferred without notice, after which this version may no longer be accessible.”
\maketitle

\section{Derivation of the stochastic mimicking of inverse Hessian matrix $-\boldsymbol{\Tilde{G}}$}
The evolution of state $X_t$ parameterized over time $t$ is governed by the
following process dynamics' stochastic differential equation (SDE):
\begin{equation}{\label{e1}}
\begin{aligned}
& dX_t=\boldsymbol{R}_t dB_t
\end{aligned}
\end{equation}
where, $\boldsymbol{R}_t$ is the diffusion coefficient, and  $B_t$ is a standard Brownian motion.\\
The process dynamics is constrained by another measurement SDE given by:
\begin{equation}{\label{e2}}
\begin{aligned}
& dY_t=\nabla f dt + dW_t
\end{aligned}
\end{equation}
where $Y_t$ is the zero-mean observation process, $\nabla f_{X_t}(:=\nabla f(X_t))$ is the gradient of cost function 
$f(X_t):\mathbb{R}^N \rightarrow \mathbb{R}^+$, and 
$W_t$  is a measurement noise (independent of process noise).
Let $X'_{t+1}$ denote the true state while $X_{t+1|t}$ represents the predicted state at $t+1$. Therefore,
\begin{equation}
    \begin{aligned}
        Y_{t+1}=\nabla f_{X'_{t+1}}+W_{t+1}
    \end{aligned}
\end{equation}
The state update obtained using the predicted state $X_{t+1|t}$ can be expressed as:
\begin{equation}
    \begin{aligned}
        \hat{X}_{t+1}=X_{t+1|t}-\boldsymbol{\Tilde{G}}(Y_{t+1}-\nabla f_{X_{t+1|t}})
    \end{aligned}
\end{equation}
It must be noted that on comparison with the Newton's-like update, the gain matrix $\boldsymbol{\Tilde{G}}$ resembles the negative Hessian-inverse (i.e. $\boldsymbol{\Tilde{G}}=-\boldsymbol{H}^{-1}$). Therefore, the error can be formally presented as $e=X'_{t+1}-\hat{X}_{t+1}$. For simplicity, we avoid the subscript $t+1$ in later part of the derivation. Accordingly, the error covariance matrix $\boldsymbol{P}$ can be written as:
\begin{equation}
    \begin{aligned}
        \boldsymbol{P}=\;&\mathbb{E}[(e-\bar{e})(e-\bar{e})^T]\\
        \implies \boldsymbol{P}=\;&\mathbb{E}[((X'-X-\overline{(X'-X)})+\boldsymbol{\Tilde{G}}((\nabla f_{X'}-\nabla f_{X})-\overline{(\nabla f_{X'}-\nabla f_{X})})+\boldsymbol{\Tilde{G}}W)\\&((X'-X-\overline{(X'-X)})+\boldsymbol{\Tilde{G}}((\nabla f_{X'}-\nabla f_{X})-\overline{(\nabla f_{X'}-\nabla f_{X})})+\boldsymbol{\Tilde{G}}W)^T]\\
        \implies \boldsymbol{P}=\;&\mathbb{E}[((X-\overline{X})+\boldsymbol{\Tilde{G}}(\nabla f_{X}-\overline{\nabla f_{X}})-\boldsymbol{\Tilde{G}}W)((X-\overline{X})+\boldsymbol{\Tilde{G}}(\nabla f_{X}-\overline{\nabla f_{X}})-\boldsymbol{\Tilde{G}}W)^T]\\
        \implies \boldsymbol{P}=\;&\mathbb{E}[(A+\boldsymbol{\Tilde{G}}B)(A+\boldsymbol{\Tilde{G}}B)^T]\\
        \implies \boldsymbol{P}=\;&\mathbb{E}[AA^T+AB^T\boldsymbol{\Tilde{G}}^T+\boldsymbol{\Tilde{G}}BA^T+\boldsymbol{\Tilde{G}}BB^T\boldsymbol{\Tilde{G}}^T]
    \end{aligned}
\end{equation}
where, $A=(X-\bar{X})$ and $B=(\nabla f_X-\overline{\nabla f_{X}}-W)$.
Using the linearity of expectation, we can split this into separate expectations:
\begin{equation}
    \begin{aligned}
        \boldsymbol{P}=&\mathbb{E}[AA^T]+\mathbb{E}[AB^T\boldsymbol{\Tilde{G}}^T]+\mathbb{E}[\boldsymbol{\Tilde{G}}BA^T]+\mathbb{E}[\boldsymbol{\Tilde{G}}BB^T\boldsymbol{\Tilde{G}}^T]\\
    \end{aligned}
\end{equation}
To obtain the expression of $\boldsymbol{\Tilde{G}}$, we minimize the trace of error covariance matrix with respect to $\boldsymbol{\Tilde{G}}$:
\begin{equation}
    \begin{aligned}
        tr(\boldsymbol{P})=tr(C)+tr(D\boldsymbol{\Tilde{G}}^T)+tr(\boldsymbol{\Tilde{G}}D^T)+tr(\boldsymbol{\Tilde{G}}E\boldsymbol{\Tilde{G}}^T)
    \end{aligned}
\end{equation}
where, $C=\mathbb{E}[AA^T]$, $D=\mathbb{E}[AB^T]$ and $E=\mathbb{E}[BB^T]$. Thus, we get,
\begin{equation}
    \begin{aligned}
        \frac{\partial tr(\boldsymbol{P})}{\partial \boldsymbol{\Tilde{G}}}=2D+2\boldsymbol{\Tilde{G}}E
    \end{aligned}
\end{equation}
Setting $ \frac{\partial tr(\boldsymbol{P})}{\partial \boldsymbol{\Tilde{G}}}=0$ gives,
\begin{equation}
    \begin{aligned}
        \boldsymbol{\Tilde{G}}=-DE^{-1}
    \end{aligned}
\end{equation}
So, the optimal $\boldsymbol{\Tilde{G}}$ that minimizes the $tr(\boldsymbol{P})$ is:
\begin{equation}
    \begin{aligned}
        \boldsymbol{\Tilde{G}}=-\mathbb{E}[AB^T](\mathbb{E}[BB^T])^{-1}
    \end{aligned}
\end{equation}
Since, $W$ is zero-mean and independent of $X$ and $\nabla f$, i.e. $\mathbb{E}[(X-\bar{X})W^T]=\mathbb{E}(X-\bar{X})\mathbb{E}[W^T]=\boldsymbol{0}$,
\begin{equation}
    \begin{aligned}
            \mathbb{E}[AB^T]=\mathbb{E}[(X-\bar{X})(\nabla f_X-\overline{\nabla f_{X}})^T]
    \end{aligned}
\end{equation}
Accordingly, 
\begin{equation}
    \begin{aligned}
        \mathbb{E}[BB^T]&=\mathbb{E}[(\nabla f_X-\overline{\nabla f_{X}})(\nabla f_X-\overline{\nabla f_{X}})^T]+\mathbb{E}[WW^T]\\
        \implies\mathbb{E}[BB^T]&=\mathbb{E}[(\nabla f_X-\overline{\nabla f_{X}})(\nabla f_X-\overline{\nabla f_{X}})^T]+\boldsymbol{Q}        
    \end{aligned}
\end{equation}
where, $\boldsymbol{Q}=\mathbb{E}[WW^T]$ represents the measurement noise covariance matrix. Therefore, the expression for $\boldsymbol{\Tilde{G}}$ can be written as:
\begin{equation}
    \begin{aligned}
        \boldsymbol{\Tilde{G}}=-\mathbb{E}[(X-\bar{X})(\nabla f_X-\overline{\nabla f_{X}})^T]\left(\mathbb{E}[(\nabla f_X-\overline{\nabla f_{X}})(\nabla f_X-\overline{\nabla f_{X}})^T]+\boldsymbol{Q} \right)^{-1}
    \end{aligned}
\end{equation} 
\begin{equation}
    \begin{aligned}
        \implies\boldsymbol{\Tilde{G}}&=-\left[\int_\Omega \left(X-\Bar{X}\right)\left(\nabla f-\overline{\nabla f}\right)^Td\mathbb{P}\right]\left[\int_\Omega  \left(\nabla f-\overline{\nabla f}\right)\left(\nabla f-\overline{\nabla f}\right)^Td\mathbb{P} + \boldsymbol{Q}\right]^{-1}
    \end{aligned}
\end{equation}

\section{Walkthrough Example: FINDER algorithm}
Consider a 5 dimensional sphere function
\begin{equation}
    f(x_1,x_2,x_3,x_4,x_5) = x_1^2 + x_2^2+x_3^2+x_4^2+x_5^2
\end{equation}
with (1,1,1,1,1) as the initial guess point.
\newline
\newline
\textbf{Step 1 - Generate Ensemble}
\lstset{language=Python,
xleftmargin=1.5cm}
\begin{lstlisting}
import torch

N = 5   # dim of design space
p = 5   # number of particles

X0 = torch.ones((N,1))
Z = torch.empty(p-1,N).uniform_(-1,1)
R = 0.1 * torch.ones(1,N)

X = torch.cat([X0, X0 + (R*Z).T], dim = 1)   # ensemble matrix

print("ensemble matrix:\n",X)

Output:

ensemble matrix : 
 tensor([[1.0000, 0.9774, 1.0538, 1.0130, 1.0964],
        [1.0000, 1.0525, 0.9924, 1.0226, 0.9661],
        [1.0000, 0.9111, 1.0330, 1.0314, 0.9254],
        [1.0000, 0.9699, 1.0046, 0.9095, 0.9455],
        [1.0000, 0.9996, 0.9198, 1.0840, 1.0809]])
\end{lstlisting}

Thus we obtain ensemble matrix $\boldsymbol{X}$.
\newline
\newline
\textbf{Step 2 - Evaluate Gradients}
\lstset{language=Python,
xleftmargin=1.5cm}
\begin{lstlisting}
def func(x):
    """
    accepts ensemble matrix
    returns loss vector
    """
    return x.pow(2).sum(dim = 1)

X.requires_grad = True
f = func(X)
f.backward(torch.ones_like(f))
G = X.grad # gradients matrix

print("loss vector:\n", f)
print()
print("gradients matrix:\n", G)

Output:

loss vector:
 tensor([5.2940, 5.0716, 4.8174, 4.6710, 5.1885], grad_fn=<SumBackward1>)

gradients matrix:
 tensor([[2.0000, 1.9548, 2.1076, 2.0260, 2.1928],
        [2.0000, 2.1049, 1.9847, 2.0453, 1.9321],
        [2.0000, 1.8222, 2.0659, 2.0629, 1.8508],
        [2.0000, 1.9398, 2.0092, 1.8190, 1.8910],
        [2.0000, 1.9991, 1.8396, 2.1680, 2.1618]])
\end{lstlisting}

Next, we sort the loss vector $f$
\begin{lstlisting}
sorted_loss, sorting_index = torch.sort(f)
print(sorted_loss)
print(sorting_index)

Output:

tensor([4.6710, 4.8174, 5.0716, 5.1885, 5.2940], grad_fn=<SortBackward0>)
tensor([3, 2, 1, 4, 0])
\end{lstlisting}

Accordingly rearrange columns of $\boldsymbol{X}$ and $\boldsymbol{G}$ in increasing order of loss vector $f$
\begin{lstlisting}
Xs = X[:,sorting_index]
Gs = G[:,sorting_index]

print(Xs)
print()
print(Gs)

Output:

tensor([[1.0130, 1.0538, 0.9774, 1.0964, 1.0000],
        [1.0226, 0.9924, 1.0525, 0.9661, 1.0000],
        [1.0314, 1.0330, 0.9111, 0.9254, 1.0000],
        [0.9095, 1.0046, 0.9699, 0.9455, 1.0000],
        [1.0840, 0.9198, 0.9996, 1.0809, 1.0000]], grad_fn=<IndexBackward0>)

tensor([[2.0260, 2.1076, 1.9548, 2.1928, 2.0000],
        [2.0453, 1.9847, 2.1049, 1.9321, 2.0000],
        [2.0629, 2.0659, 1.8222, 1.8508, 2.0000],
        [1.8190, 2.0092, 1.9398, 1.8910, 2.0000],
        [2.1680, 1.8396, 1.9991, 2.1618, 2.0000]])
\end{lstlisting}

Thus we obtain sorted ensemble matrix $\boldsymbol{X}^s$ and sorted gradient matrix $\boldsymbol{G}^s$
\newline
\newline
\textbf{Step 3 - Mimicked Inverse Hessian}
\lstset{language=Python,
xleftmargin=1.5cm}
\begin{lstlisting}
with torch.no_grad():
    mean_Xs = torch.mean(Xs, dim = 1).view(-1,1)
    mean_Gs = torch.mean(Gs, dim = 1).view(-1,1)
    
    numerator = torch.sum((Xs - mean_Xs)*(Gs - mean_Gs), dim = 1)
    denominator = torch.sum((Gs - mean_Gs)*(Gs - mean_Gs), dim = 1)
    
    B = numerator / denominator
    gamma = 1
    B = torch.nan_to_num(B)
    torch.clamp(B,0)
    B_ = B.pow(gamma)

print(B_)

Output:

tensor([0.5000, 0.5000, 0.5000, 0.5000, 0.5000])
\end{lstlisting}

**Note that the output of $B$ exactly matches the diagonals of the true inverse hessian for the sphere function. One can increase the dimension of the problem while keeping the number of particles same to verify that the expression for $B$ yields the inverse hessian diagonals.
\newline
\newline
\textbf{Step 4 - Update Ensemble}
\begin{lstlisting}
increment = torch.zeros_like(X)   # Comment this in future iterations
theta = 0.9
increment = theta * increment + B_*Gs
print(increment)

Output:

tensor([[1.0130, 1.0538, 0.9774, 1.0964, 1.0000],
        [1.0226, 0.9924, 1.0525, 0.9661, 1.0000],
        [1.0314, 1.0330, 0.9111, 0.9254, 1.0000],
        [0.9095, 1.0046, 0.9699, 0.9455, 1.0000],
        [1.0840, 0.9198, 0.9996, 1.0809, 1.0000]])

def backtracking_line_search(Xs, increment, Gs, f):
    step = 0.1
    c_alpha = 0.01
    descent = - increment[:,0:1]    # direction of descent
    delta = 1.0
    g = Gs[:,0:1]
    min_drop = c_alpha * (descent * g).sum()  # inner product
    while delta > 1e-6:
        new_x = Xs[:,0:1] - delta * descent
        loss_new = func(new_x)
        min_drop *= delta
        armijo = loss_new[0] - f.min() + min_drop
        if armijo < 0:
            step = delta
            break
        else:
            delta *= 0.5
    return step
alpha = backtracking_line_search(Xs, increment, Gs, f)
print(alpha)

Output:

1.0

X_hat = Xs - alpha * increment

print(X_hat)

Output:

tensor([[0., 0., 0., 0., 0.],
        [0., 0., 0., 0., 0.],
        [0., 0., 0., 0., 0.],
        [0., 0., 0., 0., 0.],
        [0., 0., 0., 0., 0.]], grad_fn=<SubBackward0>)
\end{lstlisting}

Concatenate best particle of ensemble matrix $\boldsymbol{X}$,i.e., $\boldsymbol({X}^{s})^1$ to $\hat{X}$ to obtain new ensemble $\boldsymbol{X}^{'s}$
\begin{lstlisting}
X_new = torch.cat([X_hat, Xs[:,0:1]], dim = 1)
f_new = func(X_new)
X0 = X_new[:,torch.argmin(f_new)] # initial point for next iteration

print(X0)

Output:

tensor([0., 0., 0., 0., 0.], grad_fn=<SelectBackward0>)
\end{lstlisting}
\underline{which is the minima for the sphere function.} 

Reader may modify the dimension of the problem and number of particles to play around while following this code sequence provided here.

\section{Burgers problem}
\begin{equation}\label{burgers}
    \begin{aligned}
        u_{,t} +uu_{,x}&=\frac{0.01}{\pi}u_{,xx}\\ 
        u(0,x)&=-\sin(\pi x)\rightarrow\text{IC}\\
        u(t,-1)&=u(t,1)=0\rightarrow\text{BCs}\\
    \end{aligned}
\end{equation}
where $x \in [-1,1]$ and $t \in [0,1]$. The PINN formulation of Eq. \eqref{burgers} is given as: 
\begin{equation}
    \begin{aligned}
        u &\leftarrow\mathbf{DNN}(t,x; \Theta)\\
        \hat{u} &= t(x^2 - 1)u - sin(\pi x)\\
        loss &= \frac{1}{N_c}\sum_{i=1}^{N_c} (\hat{u}^{(i)}_{,t} +\hat{u}^{(i)}\hat{u}^{(i)}_{,x}-\frac{0.01}{\pi}\hat{u}^{(i)}_{,xx})^2\\
        N_c &= 10000
    \end{aligned}
\end{equation}
where $\Theta$ denotes trainable parameters of the DNN.

\section{2d elasticity problem}
The governing PDEs, constitutive laws, and kinematic relations are as follows: 
\begin{equation}\label{elasticity}
    \begin{aligned}
        \text{Momentum balance laws: }&\sigma_{xx,x}+\sigma_{xy,y}+b_{x}=0\\        &\sigma_{yx,x}+\sigma_{yy,y}+b_{y}=0\\        
        &\sigma_{xy}=\sigma_{yx}\\
        \text{Constitutive laws: }&\sigma_{xx}=(\lambda+ 2\mu)\varepsilon_{xx}+\lambda\varepsilon_{yy}\\
        &\sigma_{yy}=(\lambda+ 2\mu)\varepsilon_{yy}+\lambda\varepsilon_{xx}\\
        &\sigma_{xy} = 2\mu\varepsilon_{xy}\\
        \text{Kinematic relations: }&\varepsilon_{xx}=u_{,x};\varepsilon_{yy}=v_{,y}\\
        &\varepsilon_{xy} = \frac{1}{2}\left(u_{,y}+v_{,x}\right)\\
        b_{x}=\lambda(4\pi^2\cos(2\pi x)\sin(\pi y)&-\pi \cos(\pi x)\mathcal{Q}y^3)\\
        +\mu(9\pi^2cos(2\pi x)sin(\pi y)&-\pi cos(\pi x)\mathcal{Q}y^3)\\
        b_{y}=\lambda(-3sin(\pi x)\mathcal{Q}y^2+2\pi&^2sin(2\pi x)cos(\pi y))\\
        +\mu(-6sin(\pi x)\mathcal{Q}y^2+2\pi&^2sin(2\pi x)cos(\pi y))\\
        +\mu\pi^2sin(\pi x)\mathcal{Q}y^4/4\;\;\;\;\;\;\;&
    \end{aligned}
\end{equation}
where $\sigma$ denotes the Cauchy stress tensor, $b$ the resultant body force, $u,v$ displacements along $x,y$ directions and $\varepsilon$ the infinitesimal strain tensor. We also take $\lambda=1$, $\mu=0.5$ and $\mathcal{Q}=4$. 
The PINN formulation of Eqn \eqref{elasticity} is given in the supplementary material. 
\begin{equation}\label{pinn_elasticity}
    \begin{aligned}
        u,v, \sigma_{xx}, &\sigma_{yy}, \sigma_{xy}\leftarrow\mathbf{DNNs}(x,y;\Theta)\\
        \hat{u} &= y(1-y)u\\
        \hat{v} &= x(1-x)yv\\
        \hat{\sigma}_{xx} &= x(1-x)\sigma_{xx}\\
        \hat{\sigma}_{yy} &= (1-y)\sigma_{yy} + (\lambda + 2\mu)Q sin(\pi x)\\
        loss &= \frac{1}{N_c}\sum_{i=1}^{N_c}(\hat{\sigma}^{(i)}_{xx,x} + \hat{\sigma}^{(i)}_{xy,y} + b^{(i)}_x)^2\\
        &+ (\hat{\sigma}^{(i)}_{yx,x} + \hat{\sigma}^{(i)}_{yy,y} + b^{(i)}_y)^2\\
        &+ (\hat{\sigma}^{(i)}_{xx} - (\lambda + 2\mu)\hat{u}^{(i)}_{,x} - \lambda\hat{v}^{(i)}_{,y})^2\\
        &+ (\hat{\sigma}^{(i)}_{yy} - (\lambda + 2\mu)\hat{v}^{(i)}_{,y} - \lambda\hat{u}^{(i)}_{,x})^2\\
        &+ (\sigma^{(i)}_{xy} - \mu(\hat{u}^{(i)}_{,y} + \hat{v}^{(i)}_{,x})^2\\
    \end{aligned}
\end{equation}
where, $\Theta$ denotes trainable network parameters. 

\section{Strain Gradient plasticity}
\label{sec:sgp}
The governing equations and ICs/BCs are as follows:
\begin{equation}
    \begin{aligned}
        \text{Macroforce balance: }&\tau_{,y} = 0\\
        \text{Microforce balance: }&\tau = \tau^p - k^p_{,y}\\
        \text{Constitutive law: }&\tau = \mu(u_{,y} - \gamma^p)\\ \text{Microstress: }&\tau^p = S_0\left(\frac{d^p}{d_0}\right)^m\frac{\dot{\gamma}^p}{d^p}\\
        \text{Gradient microstress: }&k^p = SL^2\gamma^p_{,y} + S_0 l^2\left(\frac{d^p}{d_0}\right)^m\frac{\dot{\gamma}^p_{,y}}{d^p}\\
        \text{Resistance to plastic flow: }&\dot{S} = H(S)d^p\text{ ; }S(y,0) = S_0\\
        \text{Effective flow rate: }&d^p = \sqrt{\mid\dot{\gamma}^p\mid^2 + l^2\mid\dot{\gamma}^p_{,y}\mid^2}\\
        \text{Imposed shear strain: }&E(t) = \frac{u^{\text{\cross[0.4pt]}}(t)}{h}\\
        \text{Displacement BCs: }&u(0,t) = 0;\; u(h,t) = u^{\text{\cross[0.4pt]}}(t)\\
        \text{Plastic strain rate BCs: }&\dot{\gamma}^p(0,t) = \dot{\gamma}^p(h,t) = 0\\
        \text{Displacement ICs: }&u(y,0) = 0\\
        \text{Plastic strain ICs: }&\gamma^p(y,0) = 0\\
    \end{aligned}
\end{equation}
where $\tau$ denotes shear stress and $y\in [0,h]$. The problem involves a long (height $h$ along $y$), thin body (infinite in $x$ and $z$) undergoing plane-strain shearing along $x$, so that the displacement $u:=u(y,t)$. We use the initial yield strength $S_0 = 100$ MPa, shear modulus $\mu = 100$ GPa, hardening/ softening function $H = 0$, reference flow rate $d_0 = 0.1$, rate sensitivity parameter $m = 0.02$, energetic length scale $L = 10$, dissipative length scale $l = 0$ and $h = 10$.